\documentclass[10pt,twocolumn,letterpaper]{article}

\usepackage{cvpr}              
\usepackage{amsmath}
\usepackage{amsfonts}
\usepackage{graphicx}
\usepackage{amssymb}
\usepackage{arydshln}
\usepackage[table]{xcolor}  
\usepackage{multirow}
\usepackage{graphicx}
\usepackage{wrapfig}
\usepackage{lipsum}
\usepackage{comment}
\usepackage{amsthm}
\usepackage{ulem}
\usepackage{mathtools}
\usepackage{booktabs,multirow,xcolor}
\usepackage[ruled,vlined,linesnumbered]{algorithm2e}
\newcommand{\gsub}[1]{\ifmmode_{\color{ForestGreen}{#1}}\else\textsubscript{\textcolor{ForestGreen}{#1}}\fi}
\newcommand{\rsub}[1]{\ifmmode_{\color{BrickRed}{#1}}\else\textsubscript{\textcolor{BrickRed}{#1}}\fi}

\usepackage{tikz}
\usetikzlibrary{arrows.meta,positioning,fit,shapes.geometric,calc}

\definecolor{cvprblue}{rgb}{0.21,0.49,0.74}
\usepackage[pagebackref,breaklinks,colorlinks,allcolors=cvprblue]{hyperref}

\def\paperID{} 
\def\confName{CVPR}
\def\confYear{2026}

\title{LATA: Laplacian-Assisted Transductive Adaptation for Conformal Uncertainty in Medical VLMs}

\author{
\begin{tabular}{c}
Behzad Bozorgtabar$^{1}$ \quad Dwarikanath Mahapatra$^{2}$ \quad Sudipta Roy$^{3}$\\
Muzammal Naseer$^{2}$ \quad Imran Razzak$^{4}$ \quad Zongyuan Ge$^{5}$\\[0.25em]
$^{1}$Aarhus University (\textcolor{cvprblue}{\textbf{A3 Lab}})
 \quad $^{2}$Khalifa University \quad $^{3}$Jio Institute\\
$^{4}$MBZUAI \quad $^{5}$Monash University\\[0.25em]
{\tt\small \{behzad\}@ece.au.dk}
\end{tabular}
}

\begin{document}
\maketitle
\begin{abstract}
Medical vision--language models (VLMs) are strong zero-shot recognizers for medical imaging, but their reliability under domain shift hinges on calibrated uncertainty with guarantees. Split conformal prediction (SCP) offers finite-sample coverage, yet prediction sets often become large (low efficiency) and class-wise coverage unbalanced—high class-conditioned coverage gap (CCV), especially in few-shot, imbalanced regimes; moreover, naively adapting to calibration labels breaks exchangeability and voids guarantees. We propose \texttt{\textbf{LATA}} (Laplacian-Assisted Transductive Adaptation), a \textit{training- and label-free} refinement that operates on the joint calibration and test pool by smoothing zero-shot probabilities over an image–image $k$NN graph using a small number of CCCP mean-field updates, preserving SCP validity via a deterministic transform. We further introduce a \textit{failure-aware} conformal score that plugs into the vision--language uncertainty (ViLU) framework, providing instance-level difficulty and label plausibility to improve prediction set efficiency and class-wise balance at fixed coverage. \texttt{\textbf{LATA}} is black-box (no VLM updates), compute-light (windowed transduction, no backprop), and includes an optional prior knob that can run strictly label-free or, if desired, in a label-informed variant using calibration marginals once. Across \textbf{three} medical VLMs and \textbf{nine} downstream tasks, \texttt{\textbf{LATA}} consistently reduces set size and CCV while matching or tightening target coverage, outperforming prior transductive baselines and narrowing the gap to label-using methods, while using far less compute. Comprehensive ablations and qualitative analyses show that \texttt{\textbf{LATA}} sharpens zero-shot predictions without compromising exchangeability.
\end{abstract}

\section{Introduction}
\label{sec:intro}
Foundation-scale vision--language models (VLMs) such as CLIP~\cite{radford2021learning} have emerged as strong zero-shot recognizers across many visual domains, including healthcare. Modality-specialized variants (e.g., radiology~\cite{wang2022medclip}, computational pathology~\cite{lu2024visual,huang2023visual}, ophthalmology~\cite{silva2025foundation}) inherit broad transfer from generalist pretraining while exposing practical failure modes when classes are fine-grained, label marginals are imbalanced, or domain frequencies at pretraining diverge from deployment. In safety–critical settings, the central question is therefore not only \textit{how accurate} a model is, but \textit{how reliably it communicates uncertainty}---ideally with guarantees that hold at deployment time.

\noindent\textbf{Conformal background.}
\textit{Conformal prediction} (CP)~\cite{vovk2012conditional,gammerman2013learning,vovk2005algorithmic} is a model-agnostic framework that wraps around any predictive model—treated as a black box—to produce set-valued outputs with finite-sample \textit{marginal} coverage guarantees. Its inductive variant, \textit{split conformal prediction} (SCP)~\cite{lei2018distribution,papadopoulos2002inductive,vovk2005algorithmic}, computes a threshold on \textit{nonconformity} scores using a held-out calibration set. Provided the calibration and test samples are \textit{exchangeable}~\cite{vovk2005algorithmic}, SCP produces prediction sets that include the correct label with a guaranteed confidence level across the test distribution. Because SCP relies solely on the model’s nonconformity scores, it applies equally well to zero-shot and lightly adapted VLMs, offering finite-sample coverage without retraining. However, for medical VLMs, two persistent challenges remain: (i) \textit{efficiency and fairness}—prediction sets often become overly large or imbalanced across classes (as measured by class-conditioned coverage gap (CCV)~\cite{ding2023class}), especially under distribution shift or when calibration data is limited and skewed; and (ii) \textit{underutilized multimodal signals}—standard nonconformity scores such as LAC~\cite{sadinle2019least}, APS~\cite{romano2020classification}, and RAPS~\cite{angelopoulos2020uncertainty} overlook informative image--text cues that correlate with failure cases and label plausibility.


\begin{figure*}[t]
    \centering
    \includegraphics[width=\textwidth]{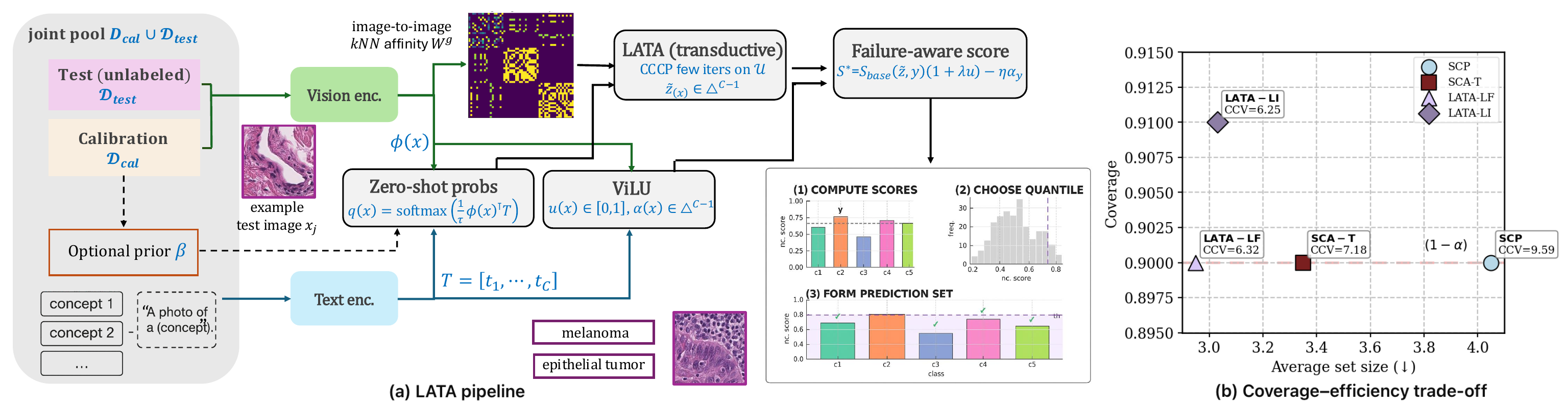}
    \caption{\textbf{\texttt{\textbf{LATA}} pipeline and coverage–efficiency trade-off.}
(a) \textbf{\texttt{\textbf{LATA}} pipeline.} Frozen vision/text encoders yield zero-shot scores $q(x)$, optionally adjusted via calibration-informed priors. \texttt{\textbf{LATA}} then refines predictions on the joint unlabeled pool $\mathcal{U}$ using a sparse $k$NN graph and CCCP updates, producing $\tilde z(x)$. A frozen ViLU module estimates difficulty $u(x)$ and attention $\alpha(x)$, forming a failure-aware score $S^\star$, which is conformalized into calibrated prediction sets.
(b) \textbf{Coverage–efficiency frontier} ($\alpha{=}0.10$, APS). \texttt{\textbf{LATA-LF}} ($\beta{=}0$) achieves SCP-level coverage with lower set size and CCV. \texttt{\textbf{LATA-LI}} ($\beta{=}0.2$) improves coverage further with minimal cost, outperforming SCA-T in both efficiency and balance.
}
    \label{fig:fig1}
    \vspace{-0.3cm}
\end{figure*}
\noindent\textbf{Adapting without breaking validity.}
In practice, one might try to narrow the domain gap by fitting a linear probe/adapter on the few calibration labels and then conformalize on the \textit{same} split—yet this \textit{double dipping} re-adjusts nonconformity scores, induces a calibration–test covariate shift, breaks exchangeability, and invalidates finite-sample coverage, even when accuracy or alignment appears to improve~\cite{silva2025trustworthy}, as illustrated in Fig.~\ref{fig:exch-corr-combined}\textbf{(b)}. \textit{This raises a central question:} 
\textbf{Can we adapt zero-shot medical VLMs to the \textit{target} distribution while \textit{retaining} split-conformal guarantees, and do so without training new task-specific parameters or consuming additional labeled subsets?} Classical full conformal prediction (FCP)~\cite{saunders1999transduction} preserves validity in a transductive manner, but its per-query, label-wise refits are prohibitive for deep VLMs. 
We thus pursue a black-box, training-free \textit{transductive} refinement that keeps SCP’s simplicity intact and avoids reusing calibration labels.

\noindent\textbf{Our remedy.}
We propose \texttt{\textbf{LATA}}, a deterministic, label- and training-free \textit{transductive} refinement that sharpens zero-shot probabilities over a shared image–image $k$-nearest neighbor ($k$NN) graph, preserving SCP validity. We also introduce a \textit{failure-aware} conformal score that reweights nonconformity using a frozen vision–language uncertainty (ViLU) module~\cite{LafonKarmim2025}, inflating scores on predicted hard inputs and deflating them for image-consistent labels. Fig.~\ref{fig:fig1} summarizes the pipeline and the empirical coverage-efficiency frontier (Fig.~\ref{fig:fig1}\textbf{(b)}). The main contributions are:

\begin{itemize}[leftmargin=1.2em,itemsep=3pt]
\item \textbf{\texttt{LATA}: label– and training–free transductive refinement.}
A deterministic mean–field update over a sparse image–image $k$NN graph built on the \textit{joint} calibration and test pool refines zero–shot posteriors. The symmetric transform preserves exchangeability and SCP validity, enabling sharper zero-shot posteriors with \textit{smaller} prediction sets and \textit{lower} CCV—achieved entirely without gradient updates or VLM finetuning.

\item \textbf{Failure-aware conformal scoring.}
We introduce a plug-in scoring mechanism that leverages a vision–language failure-prediction module~\cite{LafonKarmim2025} to \textit{reweight} nonconformity: predicted hard inputs have their nonconformity amplified, while image-consistent labels receive gentler penalties. In practice, this yields tighter sets at the target coverage with more uniform class-wise behaviour. As a controllable knob, we also allow a \textit{label-informed} prior (from calibration label marginals) computed once and applied symmetrically to calibration and test. Overall, this scoring—within \texttt{\textbf{LATA}}—approaches the set-efficiency of label-using calibrations while remaining black-box, training-free, and substantially more compute-efficient.

 \item \textbf{Comprehensive evaluation.}
Across \textit{three} medical VLMs and \textit{nine} adaptation tasks—spanning fine-grained, imbalanced, and cross-domain settings—\texttt{\textbf{LATA}} consistently reduces set size and CCV \textit{at target coverage} (LAC/APS/RAPS, $\alpha\!\in\!\{0.05,0.10\}$). Sensitivity studies show robustness to graph/solver choices and windowing, and quantify clear compute–quality trade-offs. A complexity audit against competitive baselines highlights that \texttt{\textbf{LATA}} delivers these gains with markedly lower compute and no transfer-time labels (\textbf{\textcolor{blue}{Appendix~\ref{app:complexity}}}.).

\end{itemize}

\section{Related Work}
\label{sec:related}
Prior work on reliable adaptation with VLMs spans two key directions. \textbf{\emph{Transfer learning methods}} such as prompt tuning and \textit{black-box adapters}/linear probes~\cite{yao2023visual,zhou2022learning,huang2024lp++,silva2024closer} adapt few-shot VLMs but are inductive and often break calibration–test exchangeability. Recent transductive adapters improve accuracy using test data but lack coverage guarantees~\cite{zanella2024boosting}. In parallel, \textbf{\emph{conformal prediction with VLMs}}~\cite{romano2020classification,silva2025conformal,silva2025full} extends split CP~\cite{papadopoulos2002inductive,vovk2005algorithmic} to zero-shot logits, but existing approaches typically rely on labels, assume i.i.d. conditions, or require costly refits to maintain coverage~\cite{silva2025full,silva2025conformal,silva2025trustworthy,sadinle2019least,romano2020classification,angelopoulos2020uncertainty}. Our method is transductive, label-free, and fully black-box. \textbf{For a detailed review of VLM transfer methods and conformal prediction approaches, see \textbf{\textcolor{blue}{Appendix~\ref{sec:relatedwork_supp}}}}.

\section{Method}
\label{sec:method}
We build on contrastive VLMs of the CLIP~\cite{radford2021learning} family and its modality-specialized medical variants. We begin by formalizing the zero-shot predictive distributions these models produce and briefly reviewing split conformal prediction (SCP). We then introduce a multimodal failure-prediction module, followed by our core contributions: the \texttt{\textbf{LATA}} transductive refinement procedure and a failure-aware conformal scoring scheme.

\subsection{Zero-shot VLM probabilities}
\label{subsec:zs}
Contrastive VLMs align images and text by projecting both modalities into a shared $\ell_2$-normalized embedding space. Let $\phi:\mathcal{X}\!\to\!\mathbb{R}^{D}$ be the frozen image encoder and $W=[w_1,\dots,w_C]\in\mathbb{R}^{D\times C}$ the matrix of class prototypes. For an image $x$ with feature $v=\phi(x)$ and a fixed temperature $\tau>0$, class probabilities are obtained by a temperature-scaled softmax over similarity scores:
\begin{equation}
\label{eq:1}
p_c(W,v)\;=\;\frac{\exp\!\big(v^\top w_c/\tau\big)}{\sum_{j=1}^{C}\exp\!\big(v^\top w_j/\tau\big)}\!,\qquad 
p(W,v)\in\Delta^{C-1}.
\end{equation}


\noindent\textit{Text-driven zero-shot predictions.}
In the zero-shot regime, we do not learn $W$. Instead, each prototype is fixed to the corresponding text embedding, i.e., $w_c = t_c$ for class $c$. To construct $t_c$, we encode a small set of class-specific templates (optionally including synonyms), $\ell_2$-normalize each embedding, and average them, giving $t_c = \tfrac{1}{J}\sum_{j=1}^J t_c^{(j)}$. Using these prototypes in Eq.~\eqref{eq:1} yields the standard zero-shot VLM predictive distribution. For brevity, we denote it by $q(x) = p(W,\phi(x)) \in \Delta^{C-1}$ for the base zero-shot distribution used by all subsequent components.

\subsection{Split conformal prediction (SCP)}
\label{subsec:scp}

SCP wraps a classifier with a calibration step to produce set-valued predictions at a user–chosen error level $\alpha\in(0,1)$, targeting marginal coverage $1-\alpha$. 
Let $\mathcal{D}_{\mathrm{cal}}=\{(x_i,y_i)\}_{i=1}^{n}$ be a labeled \textit{calibration} split and $\mathcal{D}_{\mathrm{test}}=\{x_j\}_{j=1}^{m}$ an unlabeled test pool, assumed \textit{exchangeable} with $\mathcal{D}_{\mathrm{cal}}$. Given a fixed \textit{nonconformity score} $S(z, y)$—which quantifies how incompatible a candidate label $y$ is with a prediction vector $z$ for input $x$—we first compute calibration scores as $s_i = S(z(x_i), y_i)$ for all $(x_i, y_i) \in \mathcal{D}_{\mathrm{cal}}$. Next, we determine the empirical $(1 - \alpha)$ quantile of the calibration scores, which serves as the threshold for set inclusion:
\begin{equation}
\label{eq:quant}
\hat{s}\;=\;\inf\!\left\{s:\;\frac{1}{n}\,\big|\{i:\; s_i\le s\}\big|\;\ge\;\frac{\lceil(n+1)(1-\alpha)\rceil}{n}\right\}.
\end{equation}

\noindent
Finally, for a new test input $x$, SCP constructs the prediction set as $\mathcal{C}(x) = \{\, y \in \{1, \dots, C\} : S(z(x), y) \le \hat{s} \,\}$, which includes all labels whose nonconformity scores fall below the calibrated threshold $\hat{s}$. Under calibration–test exchangeability and a fixed scoring rule $S$, SCP guarantees \textit{finite-sample marginal coverage}:
$\Pr\{Y\in\mathcal{C}(X)\}\ge 1-\alpha$.

\subsection{Multimodal failure prediction}
\label{subsec:vilu}
Standard conformity scores~\cite{sadinle2019least,romano2020classification,angelopoulos2020uncertainty} used in vision classification operate solely on class probabilities and therefore overlook the richer image–text structure available in VLMs. To exploit this additional signal, we employ the vision–language uncertainty (ViLU) framework~\cite{LafonKarmim2025} as a black-box module that outputs two signals for each image $x$: an instance-level failure probability $u(x) \in [0,1]$ and an image-conditioned label-attention vector $\alpha(x) \in \Delta^{C-1}$. These will later modulate the nonconformity score to improve set efficiency and class-wise balance, without changing the underlying VLM. Concretely, given the normalized image embedding $v=\phi(x)$ and the bank of normalized class text embeddings $T=[t_1,\dots,t_C]$, ViLU applies a single cross-attention layer from $v$ to $T$ to produce $\alpha(x)$ and a contextual text summary $z_t^\alpha(x)$ (see~\cite{LafonKarmim2025} for details). A small MLP $g$ then predicts the failure probability as:
\begin{equation}
\label{eq:ViLU}
u(x) = \sigma\big(g\big([\; v,\, t_{\hat{c}(x)},\, z_t^\alpha(x) \;]\big)\big),
\end{equation}
where $\hat{c}(x) = \arg\max_j q_j(x)$ is the predicted class, and $\sigma(a)=1/(1+e^{-a})$ is the logistic sigmoid. ViLU is pretrained \textit{once} on separate labeled source data to classify VLM errors using weighted binary cross-entropy, and is then frozen. At adaptation time, we apply the same fixed mapping $(u,\alpha)$ to both calibration and test examples, which preserves calibration–test exchangeability when these signals are incorporated into conformal scoring.

\subsection{\texttt{\textbf{LATA}}: A label- and training-free transductive refinement}
\label{subsec:lata}
We refine zero-shot probabilities \textit{without} touching VLM weights by adapting only the output distributions over the joint calibration and test pool. Inspired by graph \textit{Laplacian regularization}~\cite{iscen2019label,belkin2006manifold}, we propose a label-free, graph-based refinement method that is applied identically to both calibration and test samples, thereby preserving the exchangeability assumptions required by SCP. The core idea is to produce refined distributions that stay close to the initial zero-shot predictions while varying smoothly across a sparse image–image $k$NN graph.


\vspace{+0.1cm}
\noindent\textbf{Graph over the joint pool.}
Let $\mathcal{U}=\mathcal{D}_{\text{cal}}\cup\mathcal{D}_{\text{test}}$ denote the joint unlabeled pool with $N=n+m$ total samples. Let $\tilde v(x)=\phi(x)/\|\phi(x)\|_2$ be the $\ell_2$-normalized image embedding. We collect $\{\tilde v_i\}_{i=1}^N$ and build a \textit{symmetric (union)} $k$NN graph in this space, storing affinities in a sparse matrix $W^{\mathrm g}\in\mathbb{R}^{N\times N}$:
\begin{equation}
\label{eq:graph_affinity}
W^{\mathrm{g}}_{ij} =
\begin{cases}
\exp\left(-\|\tilde{v}_i - \tilde{v}_j\|_2^2 / \sigma^2 \right), & \text{if } i \sim j, \\[3pt]
0, & \text{otherwise},
\end{cases}
\end{equation}
where $i \sim j$ indicates that $i$ and $j$ are mutual $k$NN neighbors (i.e., $i \in \mathsf{kNN}(j)$ or $j \in \mathsf{kNN}(i)$), and $\sigma$ is set to the median neighbor distance unless otherwise noted. We also define the degree matrix $D = \mathrm{diag}(W^{\mathrm{g}} \mathbf{1})$ and Laplacian $L = D - W^{\mathrm{g}}$ for reference. Let $q_i = q(x_i)$ be the zero-shot probability vector for image $x_i$, and stack these into a matrix $Q = [q_1^\top; \dots; q_N^\top] \in \mathbb{R}^{N \times C}$.

\vspace{+0.2cm}
\noindent\textbf{Objective.}
We formulate \texttt{\textbf{LATA}}'s refinement objective as a regularized optimization problem over smoothed prediction vectors $\tilde z_i \in \Delta^{C-1}$, where $\Delta^{C-1}$ is the probability simplex over $C$ classes. The goal is to find refined distributions $\{\tilde z_i\in\Delta^{C-1}\}_{i=1}^N$ that (i) remain faithful to the initial zero-shot scores $q_i$, and (ii) vary smoothly across the image–image graph:
\begin{equation}
\label{eq:lata_obj}
\min_{\{\tilde z_i\in\Delta^{C-1}\}}
\underbrace{\sum_{i=1}^N \mathrm{KL}(\tilde z_i\Vert q_i)}_{\text{fidelity}}
\;+\;
\underbrace{\frac{\gamma}{2}\sum_{i,j} W^{\mathrm g}_{ij}\,\|\tilde z_i-\tilde z_j\|_2^2}_{\text{graph smoothness}},
\end{equation}
where $\gamma > 0$ controls the degree of smoothing. Expanding the quadratic gives
\(\frac{\gamma}{2}\sum_{i,j}W^{\mathrm g}_{ij}\|\tilde z_i-\tilde z_j\|_2^2
= \gamma\sum_i d_i\|\tilde z_i\|_2^2 - \gamma\sum_{i,j}W^{\mathrm g}_{ij}\tilde z_i^\top \tilde z_j\),
where $d_i=D_{ii}$ denotes the degree of node $i$ in the affinity graph. This decomposition separates the expression into a convex (quadratic norm) term and a concave (bilinear interaction) term. Following the Concave–Convex Procedure (CCCP)~\cite{yuille2001concave}, we group the convex components—including the KL fidelity term and the diagonal norm $\sum_i d_i|\tilde z_i|^2$—into the main objective, while the concave interaction term $-\gamma\sum_{i,j}W^{\mathrm g}_{ij}\tilde z_i^\top \tilde z_j$ is linearized around the current estimate. This results in an efficient fixed-point update, as shown below.

\vspace{+0.2cm}
\noindent\textbf{Efficient solver (CCCP / mean-field).}
Linearizing the concave part at iterate $t$ leads to the multiplicative update followed by row normalization:
\begin{equation}
\label{eq:lata_update}
\tilde z^{(t+1)}_{ik}\ \propto\ q_{ik}\,\exp\!\Big(\gamma\sum_{j} W^{\mathrm g}_{ij}\,\tilde z^{(t)}_{jk}\Big),
\qquad
\sum_{k=1}^{C}\tilde z^{(t+1)}_{ik}=1,
\end{equation}
initialized with $\tilde z^{(0)}=Q$ and run for $T_{\text{iter}}$ iterations (typically $5$–$10$). CCCP ensures the objective in Eq.\ \eqref{eq:lata_obj} does not increase and converges to a stationary point~\cite{yuille2001concave}. The refinement is deterministic and label-free, preserving exchangeability needed for conformal guarantees.



\vspace{+0.2cm}
\noindent\textbf{Optional prior bias.}
Let $m\in\Delta^{C-1}$ denote a class–frequency prior (e.g., Dirichlet-smoothed calibration marginals).
We inject this prior as a \textit{fixed} bias applied identically to calibration and test:
\begin{equation}
q_{ik}\ \leftarrow\ \frac{q_{ik}\,m_k^{\beta}}{\sum_{\ell=1}^{C} q_{i\ell}\,m_\ell^{\beta}}.
\label{eq:prior_bias}
\end{equation}
This is equivalent to adding a class-dependent bias to the pre-softmax logits $z_{ik}$, updating them as $z_{ik} \leftarrow z_{ik} + \beta\,\log m_k$. Here $\beta\!\in\![0,1]$ controls the strength of the bias. When $\beta{=}0$, $m_k^{0}{=}1$ for all $k$, yielding no effective prior—our default (\texttt{\textbf{LATA-LF}}). For $\beta{>}0$, we obtain a \textit{label-informed} variant (\texttt{\textbf{LATA-LI}}) that uses calibration marginals once, applied symmetrically to calibration and test, preserving exchangeability. Throughout, \texttt{\textbf{LATA (ours)}}, \texttt{\textbf{LATA}}, and \texttt{\textbf{LATA-LF}} refer to the label-free variant unless stated otherwise.

\subsection{Failure-aware conformal nonconformity}
\label{subsec:cfaps}
We enrich the conformal score with multimodal difficulty and plausibility signals from ViLU~\cite{LafonKarmim2025}. Let $u(x)\in[0,1]$ denote the (frozen) failure probability and $\alpha(x)\in\Delta^{C-1}$ the image–conditioned label–attention vector. Given class–probabilities $\tilde z(x)$ refined by \texttt{\textbf{LATA}} (Eq.~\eqref{eq:lata_update}), we define the \textit{failure-aware} nonconformity built on a standard nonconformity score (LAC/APS/RAPS):
\begin{equation}
\label{eq:cfaps}
S^\star(x,y)\;=\;S_{\text{base}}(\tilde z(x),y)\,\big(1+\lambda\,u(x)\big)\;-\;\eta\,\alpha_y(x),
\end{equation}
where $S_{\text{base}}\in\{\text{LAC},\text{APS},\text{RAPS}\}$ and $\lambda,\eta\ge 0$ are small weights. The term $u(x)$ inflates scores on hard inputs to protect coverage, while $\alpha_y(x)$ discounts labels that the image–text attention deems plausible, preventing unnecessarily large sets.



\noindent\textbf{SCP with failure-aware scoring.}
We compute calibration scores $s_i = S^\star(x_i, y_i)$ on $\mathcal{D}_{\text{cal}}$, determine the threshold $\hat s$ via Eq.~\eqref{eq:quant}, and return $\mathcal{C}(x) = {y : S^\star(x, y) \le \hat s}$ for each test input. See \textbf{\textcolor{blue}{Appendix~\ref{sec:algorithm_LATA}}} for the full \texttt{\textbf{LATA}} algorithm.

\section{Experiments}
\label{sec:experiments}

\subsection{Experimental setup}
\label{subsec:setup}

\paragraph{Medical VLMs.}
We evaluate three VLMs specialized for distinct medical imaging modalities. For \textbf{histology}, we use CONCH~\cite{lu2024visual}, which leverages a ViT-B/16 visual backbone. In \textbf{ophthalmology}, we adopt FLAIR~\cite{silva2025foundation}, while for \textbf{chest X-rays (CXR)}, we use CONVIRT~\cite{zhang2022contrastive}, pre-trained on the MIMIC-CXR dataset~\cite{johnson2019mimic}. Both FLAIR and CONVIRT share a ResNet-50 image encoder and a BioClinicalBERT~\cite{alsentzer2019publicly} text encoder.

\noindent\textbf{Datasets and tasks.}
We benchmark our models across a variety of medical imaging tasks spanning fine-grained classification, class imbalance, and domain shifts. \textbf{Histology:} We evaluate on colorectal tissue classification using \textit{NCT-CRC}~\cite{kather2018100} (fine-grained, balanced), prostate Gleason grading with \textit{SICAPv2}~\cite{silva2020going} (fine-grained), and skin lesion classification on \textit{SkinCancer}~\cite{kriegsmann2022deep} (imbalanced). \textbf{Ophthalmology:} We include diabetic retinopathy grading on \textit{MESSIDOR}~\cite{decenciere2014feedback} (fine-grained, imbalanced), myopic maculopathy staging with \textit{MMAC}~\cite{qian2024competition} (fine-grained, imbalanced), and multi-label disease classification from \textit{FIVES}~\cite{jin2022fives} (imbalanced). \textbf{CXR:} We use five diagnostic labels from \textit{CheXpert}~\cite{irvin2019chexpert} (imbalanced), fine-grained and long-tailed findings from \textit{NIH-LT}~\cite{wang2017chestx,holste2022long}, and cross-domain pneumonia classification with \textit{COVID}~\cite{chowdhury2020can,rahman2021exploring} (imbalanced, domain-shifted).

\noindent\textbf{Baselines and comparison strategies.}
We compare against state-of-the-art black-box transductive adapters and conformal wrappers: (i) \textbf{TIM}~\cite{boudiaf2020information} (gradient-based; same training setup);
(ii) \textbf{TransCLIP}~\cite{zanella2024boosting} (unsupervised GMM-based adaptation on VLM logits);
(iii) \textbf{SCA-T}~\cite{silva2025trustworthy} (entropy-minimization on the joint calibration and test pool with a marginal prior) (iv) \textbf{SCP}~\cite{angelopoulos2020uncertainty} (vanilla split conformal on zero-shot probabilities);
(v) \textbf{LinearProbe+SCP} (linear probe trained on the \textit{same} calibration labels, then wrapped with SCP—used as a negative control since it violates exchangeability.) For context, we also report \textbf{FCA}~\cite{silva2025full} as a label-aware oracle, but exclude it from black-box comparisons as it fits per-label adapters on calibration data.


\begin{table*}[t]
\centering
\setlength{\tabcolsep}{6pt}
\renewcommand{\arraystretch}{1.05}
\caption{\textbf{Conformal prediction results} with 16-shot calibration across three nonconformity scores and two error levels, $\alpha \in \{0.05, 0.10\}$. Results are averaged over tasks and modalities. ``$\downarrow$'' indicates smaller is better; \textcolor{BrickRed}{red} marks error violations. Subscripted margins denote absolute gains w.r.t.\ the SCP baseline.}
\vspace{-1mm}
\resizebox{0.78\textwidth}{!}{
\begin{tabular}{@{}c l l c ccc ccc@{}}
\toprule
& \multirow{2}{*}{\textbf{Group}} & \multirow{2}{*}{\textbf{Method}} & \multirow{2}{*}{\textbf{ACA}$\uparrow$} & \multicolumn{3}{c}{$\boldsymbol{\alpha=0.10}$} & \multicolumn{3}{c}{$\boldsymbol{\alpha=0.05}$} \\
\cmidrule(lr){5-7}\cmidrule(lr){8-10}
&  &  &  & \textbf{Cov.} & \textbf{Size}$\downarrow$ & \textbf{CCV}$\downarrow$ & \textbf{Cov.} & \textbf{Size}$\downarrow$ & \textbf{CCV}$\downarrow$ \\
\midrule

\multirow{6}{*}{\rotatebox{90}{\textbf{LAC}}}
& Baseline        & SCP                  & 50.2            & 0.890  & 3.99 & 9.96 & 0.951 & 4.88 & 5.68 \\ \cdashline{2-10}
& Labeled-Cal.  & Adapt+SCP            & $67.1\gsub{+16.9}$ & \textcolor{BrickRed}{0.842} & $2.40\gsub{-1.59}$ & $11.17\rsub{+1.21}$ & \textcolor{BrickRed}{0.921} & $3.07\gsub{-1.81}$ & $6.87\rsub{+1.19}$ \\
& Labeled-Cal.  & FCA~\cite{silva2025full}                  & $67.1\gsub{+16.9}$ & 0.896  & $2.91\gsub{-1.08}$ & $8.38\gsub{-1.58}$ & 0.952 & $3.56\gsub{-1.32}$ & $5.02\gsub{-0.66}$ \\\cdashline{2-10}
& UT            & Conf-OT~\cite{silva2025conformal}        & $53.1\gsub{+2.9}$ & 0.899  & $3.18\gsub{-0.81}$ & $9.07\gsub{-0.89}$ & 0.950 & $3.81\gsub{-1.07}$ & $5.34\gsub{-0.34}$ \\
& UT            & SCA-T~\cite{silva2025trustworthy}                 & $55.2\gsub{+5.0}$ & 0.898  & $3.30\gsub{-0.69}$ & $7.47\gsub{-2.49}$ & 0.952 & $4.03\gsub{-0.85}$ & $4.03\gsub{-1.65}$ \\

\rowcolor{cyan!20}
\cellcolor{white}& UT & \texttt{\textbf{LATA-LF}} ($\beta{=}0$) & $57.0\gsub{+6.8}$ & 0.900 & $\textbf{3.07}\gsub{-0.92}$ & $6.40\gsub{-3.56}$& 0.952 & $\textbf{3.76}\gsub{-1.12}$ & $3.35\gsub{-2.33}$ \\
\rowcolor{cyan!20}
\cellcolor{white}& UT & \texttt{\textbf{LATA-LI}} ($\beta{=}0.2$) & $\textbf{57.4}\gsub{+7.2}$
& \textbf{0.910} & $3.15\gsub{-0.84}$ & $\textbf{6.25}\gsub{-3.71}$ & \textbf{0.962} & $3.86\gsub{-1.02}$ & $\textbf{3.40}\gsub{-2.28}$\\

\midrule

\multirow{6}{*}{\rotatebox{90}{\textbf{APS}}}
& Baseline        & SCP                  & 50.2            & 0.900  & 4.05 & 9.59 & 0.952 & 4.88 & 5.54 \\ \cdashline{2-10}
& Labeled-Cal.  & Adapt+SCP            & $67.1\gsub{+16.9}$ & \textcolor{BrickRed}{0.858} & $2.56\gsub{-1.49}$ & $8.57\gsub{-1.02}$ & \textcolor{BrickRed}{0.924} & $3.19\gsub{-1.69}$ & $6.08\rsub{+0.54}$ \\
& Labeled-Cal.  & FCA~\cite{silva2025full}                  & $67.1\gsub{+16.9}$ & 0.898  & $3.06\gsub{-0.99}$ & $6.12\gsub{-3.47}$ & 0.949 & $3.67\gsub{-1.21}$ & $4.24\gsub{-1.30}$ \\
\cdashline{2-10}
& UT            & Conf-OT~\cite{silva2025conformal}        & $53.1\gsub{+2.9}$ & 0.899  & $3.13\gsub{-0.92}$ & $8.64\gsub{-0.95}$ & 0.950 & $3.63\gsub{-1.25}$ & $5.12\gsub{-0.42}$ \\
& UT            & SCA-T~\cite{silva2025trustworthy}                & $55.2\gsub{+5.0}$ & 0.900  & $3.35\gsub{-0.70}$ & $7.18\gsub{-2.41}$ & 0.954 & $4.08\gsub{-0.80}$ & $3.97\gsub{-1.57}$ \\
\rowcolor{cyan!20}
\cellcolor{white}& UT & \texttt{\textbf{LATA-LF}} ($\beta{=}0$) & $57.1\gsub{+6.9}$ & 0.900 & $\textbf{2.95}\gsub{-1.10}$ & $6.32\gsub{-3.27}$ & 0.954 & $\textbf{3.78}\gsub{-1.10}$ & $3.55\gsub{-1.99}$ \\
\rowcolor{cyan!20}
\cellcolor{white}& UT & \texttt{\textbf{LATA-LI}} ($\beta{=}0.2$) & $\textbf{57.5}\gsub{+7.3}$ & \textbf{0.910} & $3.03\gsub{-1.02}$ & $\textbf{6.25}\gsub{-3.34}$ & \textbf{0.963} & $3.88\gsub{-1.00}$ & $\textbf{3.45}\gsub{-2.09}$ \\

\midrule

\multirow{7}{*}{\rotatebox{90}{\textbf{RAPS}}}
& Baseline      & SCP                  & 50.2            & 0.901  & 4.16 & 9.55 & 0.952 & 5.12 & 5.57 \\ \cdashline{2-10}
& Labeled-Cal.  & Adapt+SCP            & $67.1\gsub{+16.9}$ & \textcolor{BrickRed}{0.856} & $2.55\gsub{-1.61}$ & $8.64\gsub{-0.91}$ & \textcolor{BrickRed}{0.923} & $3.17\gsub{-1.95}$ & $6.12\rsub{+0.55}$ \\ 
& Labeled-Cal.  & FCA~\cite{silva2025full}                  & $67.1\gsub{+16.9}$ & 0.898  & $3.05\gsub{-1.11}$ & $6.21\gsub{-3.34}$ & 0.951 & $3.66\gsub{-1.46}$ & $4.23\gsub{-1.34}$ \\
\cdashline{2-10}
& UT            & Conf-OT~\cite{silva2025conformal}        & $53.1\gsub{+2.9}$ & 0.899  & $3.42\gsub{-0.74}$ & $8.53\gsub{-1.02}$ & 0.950 & $4.09\gsub{-1.03}$ & $5.21\gsub{-0.36}$ \\
& UT            & SCA-T~\cite{silva2025trustworthy}                & $55.2\gsub{+5.0}$ & 0.903  & $3.40\gsub{-0.76}$ & $7.50\gsub{-2.05}$ & 0.954 & $4.18\gsub{-0.94}$ & $4.57\gsub{-1.0}$ \\
\rowcolor{cyan!20}
\cellcolor{white}& UT & \texttt{\textbf{LATA-LF}} ($\beta{=}0$) & $56.4\gsub{+6.2}$ & 0.903 & $\textbf{3.29}\gsub{-0.87}$ & $6.85\gsub{-2.70}$ & 0.954 & $\textbf{4.02}\gsub{-1.10}$ & $4.25\gsub{-1.32}$ \\
\rowcolor{cyan!20}
\cellcolor{white}& UT & \texttt{\textbf{LATA-LI}} ($\beta{=}0.2$) & $\textbf{56.8}\gsub{+6.6}$ & \textbf{0.913} & $3.36\gsub{-0.80}$ & $\textbf{6.75}\gsub{-2.80}$ & \textbf{0.962} & $4.10\gsub{-1.02}$ & $\textbf{4.22}\gsub{-1.35}$ \\
\bottomrule
\end{tabular}}
\vspace{-1mm}
\label{tab:cp-results}
\end{table*}


\begin{figure*}[t]
    \centering
    \includegraphics[width=\textwidth]{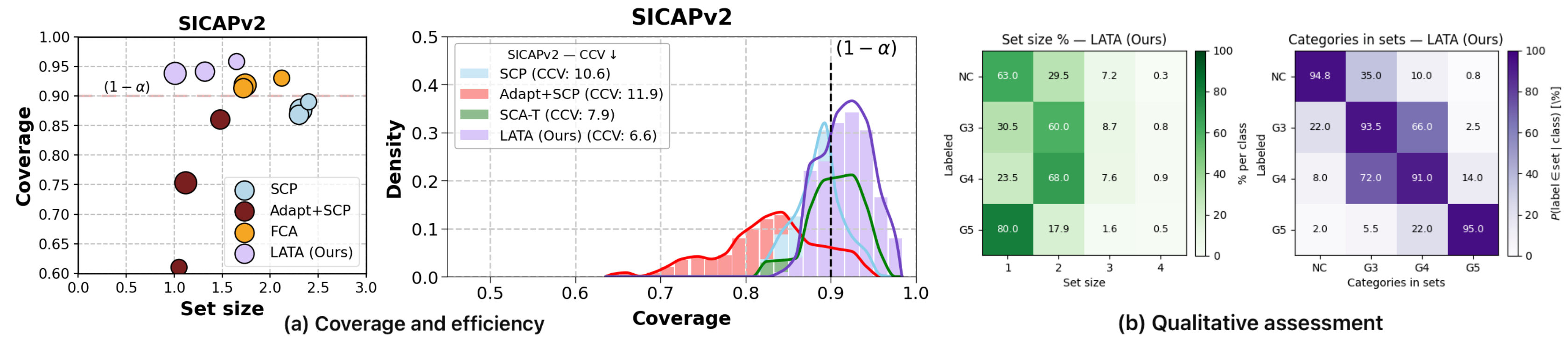}
    \caption{\textbf{SICAPv2 — coverage, efficiency, and set structure} (LAC, $\alpha{=}0.10$).  
\textbf{(a)} \textbf{Left:} Coverage–efficiency trade-off as calibration shots increase ($K\!\in\!\{4,8,16\}$; dot size encodes $K$). \texttt{\textbf{LATA (ours)}} defines the best label-free frontier—achieving smaller sets with equal or better coverage, approaching FCA without using labels at transfer.  
\textbf{Right:} Test-time coverage on $K{=}16$ splits (same seeds). Adapt+SCP under-covers; SCP/SCA-T reduce violations but remain dispersed; \texttt{\textbf{LATA}} concentrates near/above nominal with the lowest CCV. 
\textbf{(b)} Qualitative results at $K{=}16$: per-class set-size distributions (left) and label co-occurrence (right). \texttt{\textbf{LATA}} focuses uncertainty on adjacent grades (G3–G4), reducing CCV while preserving coverage.}
    \label{fig:per_dataset}
\end{figure*}


\begin{table}[t]
\centering
\setlength{\tabcolsep}{7pt}
\renewcommand{\arraystretch}{1.1}
\caption{\textbf{Transductive solvers at $\alpha{=}0.10$}. Best and second-best results are shown in \textbf{bold} and \underline{underline}, respectively. \textcolor{BrickRed}{Red} indicates violations of the target error rate.}
\resizebox{0.5\textwidth}{!}{
\begin{tabular}{@{}l c c c c c c@{}}
\toprule
\textbf{Method} & \textbf{ACA}$\uparrow$
& \textbf{Runtime $\mathbf{T}$}$\downarrow$ & \textbf{GPU}$\downarrow$ & \textbf{Cov.} & \textbf{Size}$\downarrow$ & \textbf{CCV}$\downarrow$ \\
\midrule
LAC & 50.2 & \textbf{0.00} & -- & 0.890 & 3.99 & 9.96 \\ \cdashline{1-7}[.6pt/2pt]
TIM~\cite{boudiaf2020information}           & 53.5\gsub{+3.3} & 1.12 & \textbf{0.6} & \textcolor{BrickRed}{0.888} & 3.96\gsub{-0.03} & 8.08\gsub{-1.88} \\
TransCLIP~\cite{zanella2024boosting}        & 54.8\gsub{+4.6} & 0.47 & 1.1 & \textcolor{BrickRed}{0.726} & \textbf{2.16}\gsub{-1.83} & 22.31\gsub{+12.35} \\
Conf\mbox{-}OT~\cite{silva2025conformal}    & 53.1\gsub{+2.9} & 0.60 & --  & 0.899 & 3.18\gsub{-0.81} & 9.07\gsub{-0.89} \\
SCA\mbox{-}T~\cite{silva2025trustworthy}    & 55.2\gsub{+5.0} & 1.04 & 0.6 & 0.898 & 3.30\gsub{-0.69} & 7.47\gsub{-2.49} \\
\rowcolor{cyan!20}\texttt{\textbf{LATA-LF}} ($\beta{=}0$)  & \underline{57.0}\gsub{+6.8} & \underline{0.05} & \underline{0.8} & 0.900 & \underline{3.07}\gsub{-0.92} & \underline{6.40}\gsub{-3.56} \\
\rowcolor{cyan!20}\texttt{\textbf{LATA-LI}} ($\beta{=}0.2$) & \textbf{57.4}\gsub{+7.2} & 0.05 & 0.8 & 0.910 & 3.15\gsub{-0.84} & \textbf{6.25}\gsub{-3.71} \\
\midrule
APS & 50.2 & \textbf{0.00} & -- & 0.900 & 4.05 & 9.59 \\ \cdashline{1-7}[.6pt/2pt]
TIM~\cite{boudiaf2020information}           & 53.5\gsub{+3.3} & 1.18 & \textbf{0.6} & \textcolor{BrickRed}{0.887} & 3.16\gsub{-0.89} & 7.47\gsub{-2.12} \\
TransCLIP~\cite{zanella2024boosting}        & 54.8\gsub{+4.6} & 0.40 & 1.1 & \textcolor{BrickRed}{0.733} & \textbf{2.52}\gsub{-1.53} & 21.78\gsub{+12.19} \\
Conf\mbox{-}OT~\cite{silva2025conformal}    & 53.1\gsub{+2.9} & 0.70 & --  & 0.899 & 3.13\gsub{-0.92} & 8.64\gsub{-0.95} \\
SCA\mbox{-}T~\cite{silva2025trustworthy}    & 55.2\gsub{+5.0} & 1.15 & 0.6 & 0.900 & 3.35\gsub{-0.70} &  7.18\gsub{-2.41} \\
\rowcolor{cyan!20}\texttt{\textbf{LATA-LF}} ($\beta{=}0$)  & \underline{57.1}\gsub{+6.9} & \underline{0.06} & \underline{0.8} & 0.900 & \underline{2.95}\gsub{-1.10} & \underline{6.32}\gsub{-3.27} \\
\rowcolor{cyan!20}\texttt{\textbf{LATA-LI}} ($\beta{=}0.2$) & \textbf{57.5}\gsub{+7.3} & 0.06 & 0.8 & 0.910 & 3.03\gsub{-1.02} & \textbf{6.25}\gsub{-3.34} \\
\bottomrule
\end{tabular}}
\label{tab:transductive}
\end{table}

\noindent\textbf{Implementation details.}
All experiments were run on a single GeForce RTX 4090 GPU. All methods share the same frozen vision and text encoders and identical prompts. Image features are $\ell_2$–normalized, and a \textit{symmetric} $k$NN graph is constructed via union, using Gaussian affinities with kernel bandwidth $\sigma$ set to the median neighbor distance. \texttt{\textbf{LATA}} runs in a windowed transductive setting with \textbf{$W{=}256$}, \textbf{$k{=}15$} neighbors, and \textbf{$T_{\text{iter}}{=}8$} CCCP/mean-field iterations (training-free, no gradients). We fix a single graph weight \textbf{$\gamma{=}0.35$}, selected once on a disjoint source split and reused across datasets. For large label spaces, we optionally truncate to top-$\kappa$ entries \textit{only within} \texttt{\textbf{LATA}}/ViLU (\textbf{$\kappa{=}128$}); final conformal scores always use all $C$ labels. We default to $\tau{=}1.0$ without tuning on the target domain (\textbf{\textcolor{blue}{Appendix~\ref{app:temperature}}}). The marginal prior is applied once to logits and held fixed across splits. \texttt{\textbf{LATA (ours)}} uses \textbf{$\beta{=}0$}, while \texttt{\textbf{LATA-LI}} applies \textbf{$\beta{=}0.2$} from calibration marginals. We consider standard nonconformity rules \textbf{LAC}~\cite{sadinle2019least}, \textbf{APS}~\cite{romano2020classification}, and \textbf{RAPS}~\cite{angelopoulos2020uncertainty} (defaults $k_{\text{reg}}{=}1$, $\gamma_{\text{raps}}{=}10^{-3}$). The frozen ViLU head outputs per-sample risk $u(x)$ and image-conditioned label attention $\alpha(x)$; unless otherwise stated we fix \textbf{$\lambda{=}0.5$} and \textbf{$\eta{=}0.25$} once on disjoint source data. Target error rates are $\alpha\!\in\!\{0.10,0.05\}$. Additional sensitivity analyses on the temperature parameter $\tau$, along with the full hyperparameter grids for the graph-based and failure-aware components (e.g., $k$, $T_{\text{iter}}$, $\gamma$, $\kappa$, $\beta$, $(\lambda,\eta)$), are provided in \textbf{\textcolor{blue}{Appendix~\ref{app:ablations_supp}}}.




\noindent\textbf{Evaluation protocol.}
For each task, we form a labeled calibration split of size $n_{\text{cal}} = C \times K$ with $K{=}16$ shots per class, sampled to match the task’s label marginals. We report balanced classwise accuracy (\textbf{ACA}) and standard conformal metrics~\cite{silva2025full}: marginal coverage (\textbf{Cov.}), average set size (\textbf{Size}), and class-conditioned coverage gap (\textbf{CCV})~\cite{ding2023class}. Computational efficiency is measured by peak GPU memory (\textbf{GPU}, in GB) and wall-clock inference time (\textbf{T}, in seconds).
Unsupervised transductive (\textbf{UT}) methods operate on a sliding window $\mathcal{U}_w$ of fixed size \textbf{$W{=}256$}, drawn from the union of the calibration split and the current test mini-batch. Calibration scores use refined probabilities for UT methods (zero-shot for SCP); we take the empirical $(1{-}\alpha)$ quantile and conformalize test points as in Eq.~\eqref{eq:quant}. All results are averaged over 100 seeds by re-sampling the $K$-shot calibration split. Formal definitions of nonconformity scores and evaluation metrics appear in \textbf{\textcolor{blue}{Appendix~\ref{app:scores_metrics}}}. A detailed comparison of per-window \textbf{compute cost} and access assumptions versus baselines is given in \textbf{\textcolor{blue}{Appendix~\ref{app:complexity}}}.

\subsection{Main results}
\noindent\textbf{Conformal prediction analysis.}
Table~\ref{tab:cp-results} summarizes performance across three nonconformity scores (LAC, APS, RAPS) and two target error levels ($\alpha\in\{0.10,0.05\}$). We separate \textit{label-using} methods (\emph{Labeled-Cal.}) that adapt on calibration labels (Adapt+SCP, FCA) from \emph{unsupervised transductive} (UT) methods (Conf-OT, SCA-T, \texttt{\textbf{LATA}}). Since \texttt{\textbf{LATA}} never touches calibration labels at transfer time, we treat the labeled calibration block as contextual rather than directly comparable. To ensure an apples-to-apples comparison, we focus our evaluation within the UT track.

\noindent\textbf{Does a label-free UT method improve efficiency and fairness without breaking coverage?}
Across all scores and both $\alpha$’s, \texttt{\textbf{LATA-LF}} ($\beta{=}0$; our default ``\texttt{\textbf{LATA (ours)}}’’) consistently attains nominal or higher coverage while \textit{Pareto‑dominating} strong UT baselines in both efficiency and class‑wise balance. Relative to SCA-T at $\alpha{=}0.10$, \texttt{\textbf{LATA-LF}} shrinks average set size by roughly \textbf{7–12\%} (e.g., APS: $3.35\!\to\!2.95$; LAC: $3.30\!\to\!3.07$) and reduces CCV by about \textbf{10–15\%} (e.g., LAC: $7.47\!\to\!6.40$, APS: $7.18\!\to\!6.32$), with similar gains at $\alpha{=}0.05$ (e.g., LAC size $4.03\!\to\!3.76$, CCV $4.03\!\to\!3.35$).
Compared to Conf-OT, \texttt{\textbf{LATA}} achieves comparable or smaller sets (e.g., APS size $3.13\!\to\!2.95$) but substantially lower CCV (APS $8.64\!\to\!6.32$), indicating more balanced per-class coverage under medical domain shift.
At the same time, \texttt{\textbf{LATA}} improves calibrated accuracy (ACA) over SCA-T by about \textbf{1–2.5\%}  across scores (e.g., APS: $55.2\!\to\!57.1$), suggesting that its gains are not obtained by trivially inflating prediction sets.



\noindent\textbf{How much does a weak label-informed prior help?}
Turning on the calibration-marginal prior in \texttt{\textbf{LATA-LI}} ($\beta{=}0.2$) yields a simple, \textit{fixed} knob to trade a bit of efficiency for extra coverage.
For instance, under APS at $\alpha{=}0.10$, coverage increases from $0.900$ to \textbf{0.910} while set size grows only slightly ($2.95\!\to\!3.03$) and CCV even improves ($6.32\!\to\!6.25$); analogous patterns hold for LAC and RAPS, and for $\alpha{=}0.05$.
Thus \texttt{\textbf{LATA-LI}} provides a principled way to tighten empirical coverage with minimal loss in efficiency, still preserving SCP validity since the prior is applied once and identically to calibration and test.


\begin{figure*}[t]
    \centering
    \includegraphics[width=\textwidth]{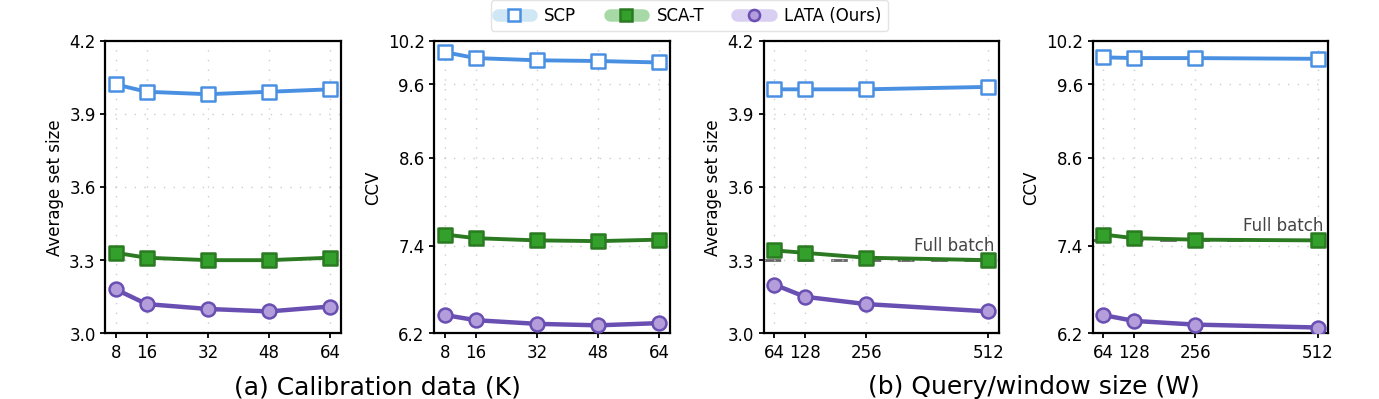}
    \caption{\textbf{Ablations on shots and window size} (LAC, $\alpha{=}0.10$; averages across tasks). 
\textbf{(a)} Effect of calibration shots $K$. 
\textbf{(b)} Effect of query/window size $W$ (dashed line marks the full-batch limit). 
Across both sweeps, \texttt{\textbf{LATA}} achieves smaller prediction sets and lower CCV than baselines while preserving nominal coverage.}
    \label{fig:fig4}
\end{figure*}

\noindent\textbf{Approaching label-using methods without labels.} 
Despite being fully label-free at transfer time, \texttt{\textbf{LATA}} approaches the behaviour of FCA, which performs full conformal adaptation on calibration labels.
Under APS at $\alpha{=}0.10$, \texttt{\textbf{LATA-LI}} attains coverage $0.910$ with size $3.03$ and CCV $6.25$, closely matching FCA (coverage $0.898$, size $3.06$, CCV $6.12$) while using no target-domain labels during adaptation.
Crucially, Adapt+SCP, which reuses calibration labels in an exchangeability-violating way, systematically under-covers (e.g., APS: $0.858$ at $\alpha{=}0.10$), whereas \texttt{\textbf{LATA}} maintains nominal coverage.
Overall, these results show that a lightweight, training-free, label-free transductive refinement can recover much of the efficiency and fairness of label-using methods, while strictly respecting the conformal prediction assumptions.

\noindent\textbf{Gleason Grading: coverage, and efficiency.}
Fig.~\ref{fig:per_dataset}\textbf{(a)} evaluates our most challenging task—Gleason grading on \textit{SICAPv2} (histology, LAC, $\alpha{=}0.10$). The \textbf{left panel} of Fig.~\ref{fig:per_dataset}\textbf{(a)} shows the coverage–efficiency trade‑off as the number of calibration shots increases ($K\in{4,8,16}$; marker size encodes $K$). While Adapt+SCP and FCA produce compact prediction sets, they fluctuate around the target coverage and rely on calibration labels. Among label‑free methods, \texttt{\textbf{LATA (ours)}} consistently defines the best top‑left frontier—achieving smaller sets than SCP and SCA‑T at comparable or slightly higher coverage, even in the low‑shot regime. This reinforces the trends seen in Table~\ref{tab:cp-results}, confirming that \texttt{\textbf{LATA}} offers strong efficiency gains without sacrificing validity. The \textbf{right panel} of Fig.~\ref{fig:per_dataset}\textbf{(a)} plots test-time coverage distributions averaged over random-seed trials on the  
$K{=}16$ calibration splits (same seeds for all methods) to compare conformal strategies under identical conditions. Reusing calibration labels for both adaptation and conformalization (Adapt+SCP) breaks exchangeability and under-covers (mass left of $(1{-}\alpha)$). SCP centers nearer the target but remains dispersed with elevated CCV; SCA-T tightens dispersion. In contrast, \texttt{\textbf{LATA (ours)}} (label-free, black-box) concentrates at/above nominal with the lowest CCV, delivering calibrated and uniformly reliable uncertainty under few-shot, fine-grained conditions. Further per-dataset coverage plots and $K$-sweeps across all nine tasks are provided in \textbf{\textcolor{blue}{Appendix~\ref{app:qualitative_supp}}}.

\noindent\textbf{Gleason Grading: qualitative assessment.}
We demonstrate the practical value of conformal prediction in medical image analysis, highlighting its utility in histopathological \textit{grading} tasks characterized by high uncertainty. Fig.~\ref{fig:per_dataset}\textbf{(b)} visualizes the \textit{prediction–set size distribution} (left) and \textit{class co-occurrence within sets} (right) for \texttt{\textbf{LATA (ours)}} on \textit{SICAPv2} at $\alpha{=}0.10$. As expected, extreme categories behave confidently: NC and G5 are mostly singletons (63\% and 80\% size-1, respectively), while the ambiguous mid-grades G3/G4 concentrate on size-2 sets (60–68\%) with only modest size-3 mass ($\sim$7–9\%). The co-occurrence heatmap shows strong diagonals (NC/G3/G5 $\ge$93\%, G4 $=72\%$ reflecting known inter-rater ambiguity) and clinically plausible adjacency: G3$\leftrightarrow$G4 co-appear frequently (66\%/91\%), whereas \textit{far} pairs are suppressed (e.g., NC$\leftrightarrow$G5 $\le$2\%). Together, these patterns explain \texttt{\textbf{LATA}}’s lower CCV and smaller sets in Table~\ref{tab:cp-results}: uncertainty concentrates on adjacent grades while suppressing spurious overlap—offering more trustworthy, efficient prediction sets.

\vspace{3pt}
\noindent\textbf{Transductive solvers.}
Table~\ref{tab:transductive} asks: \textit{Can we improve set efficiency and class-wise fairness without sacrificing coverage or incurring high compute cost?} Within the UT block, both SCA-T and \texttt{\textbf{LATA}} outperform the SCP baselines (LAC/APS), but \texttt{\textbf{LATA}} achieves the strongest results overall. The label-free variant \texttt{\textbf{LATA-LF}} ($\beta{=}0$) yields the \textit{smallest average sets} (LAC: 3.07; APS: 2.95) and the \textit{lowest class-conditioned coverage gap} (CCV: LAC 6.40; APS 6.32), indicating more balanced per-class coverage than both SCA-T and Conf-OT—all while maintaining or exceeding nominal coverage. The label-informed \texttt{\textbf{LATA-LI}} ($\beta{=}0.2$) tightens coverage (e.g., APS: \mbox{0.910}) with only a minor size increase and further CCV reduction. By contrast, TIM and TransCLIP shrink sets but \textit{under-cover}, with TransCLIP also exhibiting high CCV. Despite its improvements, \texttt{\textbf{LATA}} \textbf{remains lightweight}: the deterministic CCCP refinement over a \(k\)NN graph adds only \(\boldsymbol{\sim\!0.05\text{–}0.06}\,\mathbf{s/img}\) and \(\boldsymbol{\sim\!0.8}\,\mathbf{GB}\), and—crucially—is applied identically to calibration and test, preserving split-conformal validity while outperforming existing transductive solvers.

\subsection{Ablation studies}

\begin{figure}[t]
\hspace*{-0.5cm}
    \centering
    \includegraphics[width=0.5\textwidth,height=3.7cm]{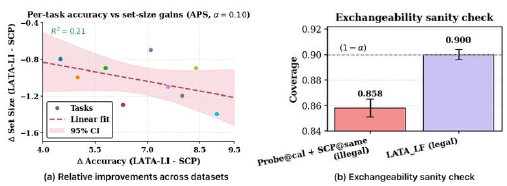}
    \vspace{-0.5cm}
    \caption{\textbf{Exchangeability and per-dataset $\Delta$accuracy–$\Delta$set-size} (APS, $\alpha{=}0.10$).
\textbf{(a)} Across datasets, \texttt{\textbf{LATA-LI}} yields $\Delta\text{Accuracy}{>}0$ and $\Delta\text{Set Size}{<}0$ vs.\ SCP, with a weak linear fit (small $R^2$), indicating efficiency gains are not merely due to accuracy increases.
\textbf{(b)} Exchangeability sanity check: the invalid \textit{Probe@cal + SCP@same} under-covers, while \texttt{\textbf{LATA-LF}} (shared, label-free transform) stays near the nominal $(1{-}\alpha)$ across random seed trials.}
    \label{fig:exch-corr-combined}
    \vspace{-0.3cm}
\end{figure}

\paragraph{Effect of calibration shots $K$ and window size $W$.}
Fig.~\ref{fig:fig4} (LAC, $\alpha{=}0.10$) explores the impact of (a) calibration shots $K$ and (b) transductive window size $W$. Across all settings, \texttt{\textbf{LATA (ours)}} consistently delivers the best trade-off—maintaining the smallest sets and lowest CCV at nominal coverage. As $K$ increases, its set size stays tight ($\mathbf{3.18}\to\mathbf{3.08}$) with CCV dropping slightly ($\mathbf{6.45}\to\mathbf{6.28}$). In contrast, SCA-T remains near $3.30$ (CCV $\approx 7.4$–$7.6$), and SCP, while valid, lags in efficiency. Increasing $W$ leads to mild gains for all UT methods. For \texttt{\textbf{LATA}}, performance improves from $\sim3.12$ to $\sim3.06$ (CCV $\sim6.3$ to $\sim6.2$) as $W$ grows from $64$ to $512$, saturating near $W{\approx}256$ (dashed). Overall, \texttt{\textbf{LATA (ours)}} remains efficiently fair, with minimal sensitivity to $K$ or $W$ and a low, stable compute footprint.

\noindent\textbf{Component analysis.}
Table~\ref{tab:ablate-lac} isolates the effect of each module in our framework. Starting from SCP (no \texttt{\textbf{LATA}}, no ViLU), adding only the risk term $u$ reduces CCV modestly with almost unchanged set size, while adding only the image-conditioned $\alpha$ shrinks sets but yields limited fairness gains.
Turning on the refinement alone---\textbf{\texttt{\textbf{LATA}} (no ViLU)}
with $\lambda{=}\eta{=}0$---delivers the bulk of the improvement at fixed coverage, cutting Size from $3.99\!\to\!3.14$ and CCV from $9.96\!\to\!6.65$ at $\alpha{=}0.10$ (similarly at $\alpha{=}0.05$: $4.88\!\to\!3.86$, $5.68\!\to\!3.60$).
Combining all components in \texttt{\textbf{LATA (ours)}} (\texttt{\textbf{LATA}}+$u$+$\alpha$) yields the best trade-off and highest ACA (Size $3.07$, CCV $6.40$ at $\alpha{=}0.10$; $3.76$ and $3.35$ at $\alpha{=}0.05$), while maintaining nominal coverage.

\begin{table}[t]
\centering
\setlength{\tabcolsep}{3.8pt}
\renewcommand{\arraystretch}{1.08}
\caption{\textbf{Component ablations (LAC)} at $\alpha\!\in\!\{0.10,0.05\}$; averages across tasks. Columns \texttt{\textbf{LATA}}/\boldmath$u$/\boldmath$\alpha$ indicate active modules. \emph{SCP}: no \texttt{\textbf{LATA}}, no ViLU. \textbf{\texttt{\textbf{LATA}} (no ViLU)}: \texttt{\textbf{LATA}} with $\lambda{=}\eta{=}0$ (both $u$ and $\alpha$ off).
\texttt{\textbf{LATA (ours)}}: \texttt{\textbf{LATA}}+$u$+$\alpha$ with $\beta{=}0$. Best and second-best in \textbf{bold}/\underline{underline}.}
\resizebox{0.49\textwidth}{!}{%
\begin{tabular}{@{}l ccc c ccc ccc@{}}
\toprule
& \multicolumn{3}{c}{\textbf{Components}} & & \multicolumn{3}{c}{$\boldsymbol{\alpha=0.10}$} & \multicolumn{3}{c}{$\boldsymbol{\alpha=0.05}$} \\
\cmidrule(lr){2-4}\cmidrule(lr){6-8}\cmidrule(lr){9-11}
\textbf{Variant} & \textbf{LATA} & \textbf{$u$} & \textbf{$\alpha$} & \textbf{ACA}$\uparrow$
& \textbf{Cov.} & \textbf{Size}$\downarrow$ & \textbf{CCV}$\downarrow$
& \textbf{Cov.} & \textbf{Size}$\downarrow$ & \textbf{CCV}$\downarrow$ \\
\midrule
\rowcolor{gray!06} SCP        & $\times$ & $\times$ & $\times$ & 50.2 & 0.890 & 3.99 & 9.96 & 0.951 & 4.88 & 5.68 \\
$u$-only                          & $\times$ & \checkmark & $\times$ & 50.6 & 0.900 & 3.95 & 9.15 & 0.952 & 4.80 & 5.35 \\
\rowcolor{gray!06} $\alpha$-only & $\times$ & $\times$ & \checkmark & 50.8 & 0.900 & 3.60 & 8.85 & 0.952 & 4.62 & 5.20 \\
\textbf{\texttt{\textbf{LATA}} (no ViLU)}                         & \checkmark & $\times$ & $\times$ & \underline{55.6} & 0.900 & \underline{3.14} & \underline{6.65} & 0.952 & \underline{3.86} & \underline{3.60} \\
\rowcolor{cyan!12}\texttt{\textbf{LATA (ours)}} & \checkmark & \checkmark & \checkmark & \textbf{57.0} & \textbf{0.900} & \textbf{3.07} & \textbf{6.40} & \textbf{0.952} & \textbf{3.76} & \textbf{3.35} \\
\bottomrule
\end{tabular}}
\label{tab:ablate-lac}
\end{table}

\noindent\textbf{Exchangeability \& accuracy–efficiency.}
Fig.~\ref{fig:exch-corr-combined}\textbf{(a)} plots, per dataset, the change relative to SCP (APS, $\alpha{=}0.10$): most points for \texttt{\textbf{LATA-LI}} lie in the desirable quadrant ($\Delta\text{Accuracy}{>}0$, $\Delta\text{Set Size}{<}0$), and the small $R^2$ of the fitted line indicates only a weak correlation—i.e., \texttt{\textbf{LATA}}’s set-size reductions are not a trivial by-product of accuracy gains.
Fig.~\ref{fig:exch-corr-combined}\textbf{(b)} demonstrates that “double-dipping’’ (training a probe on calibration labels and conformalizing on the same split) breaks exchangeability and leads to under-coverage, whereas \texttt{\textbf{LATA-LF}}, which applies a deterministic, label-free mapping identically to calibration and test, remains close to the nominal $(1{-}\alpha)$ across random seed trials.




\noindent \textbf{Additional Ablations.}
To comprehensively assess the robustness, efficiency, and practicality of our method, we conduct a series of ablation studies—covering \textbf{hyperparameter sensitivity} grids, \textbf{compute--quality trade-offs}, the impact of \textbf{ViLU domain mismatch}, and a \textbf{resource-aware gating variant} (both under LAC). Full results are provided in \textbf{\textcolor{blue}{Appendix~\ref{app:ablations_supp}}}.


\section{Conclusion}
We introduced \texttt{\textbf{LATA}}, a deterministic, training- and label-free transductive refinement for medical VLMs
that improves zero-shot predictions through KL-regularized smoothing over a sparse $k$NN graph, optimized via mean-field inference. This fixed, black-box mapping preserves split-conformal validity while producing smaller and more balanced prediction sets. Across three modality-specialized VLMs and nine medical tasks, \texttt{\textbf{LATA}} consistently reduces set size and CCV at target coverage, outperforms prior label-free transductive baselines, and narrows the gap to label-using methods at substantially lower compute.

\noindent\textbf{Limitations.}
\texttt{\textbf{LATA}}’s gains depend on embedding quality and the ViLU signals; severe domain mismatch can blunt benefits. It also assumes a modest calibration split and is instantiated for multi-class classification. Extending to dense/structured outputs, very large label spaces, and stronger shift robustness is left for future work.




\appendix
\setcounter{figure}{0}
\renewcommand{\thefigure}{S\arabic{figure}}

\setcounter{table}{0}
\renewcommand{\thetable}{S\arabic{table}}

\section{Algorithm of \texttt{\textbf{LATA}}} \label{sec:algorithm_LATA}
The end-to-end \texttt{\textbf{LATA}} procedure is summarized in Algorithm~\ref{alg:lata_scp_compact}.

\SetAlgoSkip{}

\begin{algorithm*}[t]
\small  
\caption{\texttt{\textbf{LATA}} with Failure-Aware Conformal Prediction}
\label{alg:lata_scp_compact}
\SetKwInOut{KwIn}{Inputs}
\SetKwInOut{KwOut}{Output}
\KwIn{Frozen VLM $(\phi,W)$; labeled calibration $\mathcal{D}_{\mathrm{cal}}=\{(x_i,y_i)\}_{i=1}^{n}$; unlabeled test $\mathcal{D}_{\mathrm{test}}=\{x_j\}_{j=1}^{m}$; frozen ViLU; error level $\alpha$; nonconformity rule $S_{\text{base}}\!\in\!\{\text{LAC},\text{APS},\text{RAPS}\}$; hyperparameters $(k,\sigma,\gamma,T_{\text{iter}},\lambda,\eta)$; optional prior $(\beta,m)$.}
\KwOut{Prediction sets $\{\mathcal{C}(x_j)\}_{j=1}^{m}$.}

$\mathcal{U}\leftarrow \mathcal{D}_{\mathrm{cal}}\cup\mathcal{D}_{\mathrm{test}}$ \tcp*[l]{joint pool}
\ForEach{$x\in\mathcal{U}$}{
  $q(x)\leftarrow p(W,\phi(x))$ \tcp*[l]{Eq.~\eqref{eq:1}; zero-shot probs, $q(x)\in\Delta^{C-1}$}
}

\tcp*[l]{\textbf{Block 1: \texttt{LATA} (deterministic transductive refinement)}}
Build a \textit{symmetric} $k$NN graph $W^{\mathrm g}$ on $\{\phi(x)/\|\phi(x)\|_2: x\in\mathcal{U}\}$ using union; weights via Gaussian kernel with bandwidth $\sigma$ \tcp*[l]{Sec.~\ref{subsec:lata}}
\If{$\beta>0$}{
  \ForEach{$x\in\mathcal{U}$}{
    $q(x)\leftarrow \mathrm{Renorm}\!\big(q(x)\odot m^{\beta}\big)$ \tcp*[l]{$[q_{ik}\!\leftarrow\! q_{ik}m_k^{\beta}/\sum_{\ell} q_{i\ell}m_{\ell}^{\beta}]$}
  }
}
Initialize $\tilde z^{(0)}(x)\leftarrow q(x)$ for all $x\in\mathcal{U}$\;
\For{$t=1$ \KwTo $T_{\text{iter}}$}{
  \ForEach{$x_i\in\mathcal{U}$}{
    \For{$c=1$ \KwTo $C$}{
      $m_{ic}\leftarrow \sum_{j} W^{\mathrm g}_{ij}\,\tilde z^{(t-1)}_{jc}$\;
      $\tilde z^{(t)}_{ic}\leftarrow q_{ic}\,\exp\!\big(\gamma\,m_{ic}\big)$
    }
    $\tilde z^{(t)}_{i}\leftarrow \tilde z^{(t)}_{i}\big/\sum_{c=1}^{C}\tilde z^{(t)}_{ic}$ \tcp*[l]{row-normalize}
  }
}
Set $\tilde z(x)\leftarrow \tilde z^{(T_{\text{iter}})}(x)$ for all $x\in\mathcal{U}$ \tcp*[l]{Eq.~\eqref{eq:lata_update}}

\tcp*[l]{\textbf{Block 2: Failure-aware conformal prediction}}
\ForEach{$x\in\mathcal{U}$}{
  $(u(x),\alpha(x))\leftarrow \text{ViLU}(x)$ \tcp*[l]{frozen head; identical on cal \& test}
}
\For{$i=1$ \KwTo $n$}{
  $s_i \leftarrow S_{\text{base}}(\tilde z(x_i),y_i)\,\bigl(1+\lambda\,u(x_i)\bigr)\;-\;\eta\,\alpha_{y_i}(x_i)$ \tcp*[l]{Eq.~\eqref{eq:cfaps}}
}
$\hat{s}\leftarrow$ empirical $(1-\alpha)$ quantile of $\{s_i\}_{i=1}^{n}$ \tcp*[l]{Eq.~\eqref{eq:quant}}
\ForEach{$x_j\in\mathcal{D}_{\mathrm{test}}$}{
  $\mathcal{C}(x_j)\leftarrow \bigl\{\,y\in\{1,\dots,C\}:\; S_{\text{base}}(\tilde z(x_j),y)\bigl(1+\lambda u(x_j)\bigr)-\eta\,\alpha_y(x_j)\le \hat{s}\,\bigr\}$\;
}
\Return{$\{\mathcal{C}(x_j)\}_{j=1}^{m}$}
\end{algorithm*}

\section{Related Work}
\label{sec:relatedwork_supp}
\paragraph{Transfer learning in VLMs.} Foundation VLMs deliver strong zero-shot recognition, but performance can deteriorate when downstream concepts are under-represented during pretraining, motivating few-shot transfer. Popular strategies include \textit{prompt learning} (optimizing task-specific textual tokens)~\cite{yao2023visual,zhou2022learning} and \textit{black-box adapters}/linear probes that operate on frozen embeddings; the latter often yields competitive accuracy at lower compute by blending visual and text logits (e.g., constrained or text-informed probes such as CLAP~\cite{silva2024closer} or LP++~\cite{huang2024lp++}). While effective, these methods are typically \textit{inductive}: they adapt from the labeled calibration split but do not exploit the unlabeled test distribution, and—when the same labels are reused for conformal calibration—can compromise exchangeability. Recent \textit{transductive} black-box approaches (e.g., TransCLIP~\cite{zanella2024boosting}) incorporate unlabeled test data via simple generative priors, improving accuracy but leaving coverage guarantees unaddressed. Our work departs from accuracy-only transfer by targeting \textit{reliable} few-shot adaptation: \texttt{\textbf{LATA}} is label-free, training-free, and transductive, and we couple it with a failure-aware conformal score—preserving split-conformal validity while improving set efficiency and class-wise balance at fixed coverage.

\begin{table*}[t]
\centering
\setlength{\tabcolsep}{5.5pt}
\renewcommand{\arraystretch}{1.08}
\caption{\textbf{Complexity \& compute profile}. Asymptotics in $N{=}n{+}m$, $C$, $K$, $d$.}
\resizebox{\linewidth}{!}{
\begin{tabular}{@{}l c c c l c c c@{}}
\toprule
\textbf{Method} & \textbf{Labels at transfer?} & \textbf{Training?} & \textbf{Per-query refits} & \textbf{Dominant complexity} & \textbf{Memory} & \textbf{Time/img (s)$^\dagger$} & \textbf{SCP validity} \\
\midrule
\textbf{LATA (ours)} & No & No & None &
$O(N \log N \cdot d) \;+\; O(T_{\text{iter}}\,k\,N\,C) \;+\; O(nC)$ &
$O(kN{+}NC)$ & \textbf{0.05--0.06} & Yes \\
SCA-T & No & Yes (unsup.) & None &
$O(T' N C) \;+\; O(nC)$ & logits+buffers & 1.04--1.15 & Yes \\
Conf-OT & No & No & None &
$O(S\,K\,(N{+}m)) \;+\; O(nC)$ & OT matrices & 0.60--0.70 & Yes \\
FCA & \textbf{Yes} (cal.) & \textbf{Yes} & \textbf{$O(C)$} &
$O(C\!\times\!\text{fit\_cost}) \;+\; O(nC)$ & probe heads+logits & --- (solver-dep.) & Yes \\
\bottomrule
\end{tabular}}
\vspace{1mm}
{\footnotesize $^\dagger$Measured on our setup (same backbone, prompts, hardware) and averaged over LAC/APS at $\alpha{=}0.10$.}
\label{tab:complexity_comparison}
\end{table*}

\paragraph{Conformal prediction with VLMs.}
Conformal prediction (CP)~\cite{vovk2012conditional,gammerman2013learning} provides finite-sample coverage guarantees; in vision, it is commonly used in its split form SCP~\cite{papadopoulos2002inductive,vovk2005algorithmic} with black-box models and standard nonconformity scores (LAC~\cite{sadinle2019least}, APS~\cite{romano2020classification}, RAPS~\cite{angelopoulos2020uncertainty}). However, these methods rely on models trained under the assumption that the training and test data are independently and identically distributed (i.i.d.)—an assumption that does not hold for pre-trained VLMs, which are the focus of our work. Extending CP beyond image-only classifiers to VLMs is recent: Conf-OT~\cite{silva2025conformal} rebalances zero-shot logits via optimal transport before SCP, improving size but offering limited control of class-wise balance; FCA~\cite{silva2025full} performs full-conformal adaptation with labeled calibration data. To preserve exchangeability, it evaluates test queries via per-label refits---effectively \(\mathcal{O}(C)\) gradient-based updates per query/window. This delivers strong accuracy and compact sets, but at the expense of heavy compute, higher latency, and a departure from the black-box setting; and SCA-T~\cite{silva2025trustworthy} regularizes predictions transductively on the unlabeled pool, improving coverage/efficiency without labels but without leveraging multimodal failure/plausibility cues. 
Our method differs by being \textit{label-free, training-free, and black-box}: it smooths zero-shot probabilities over an image–image $k$NN graph built on the joint calibration and test pool and augments SCP with a vision–language uncertainty head for failure-aware scoring. The transform is deterministic and applied identically to both splits (preserving SCP validity), while delivering smaller sets and lower CCV at the desired coverage.

\section{Computational Complexity}
\label{app:complexity}
In this section, we compare the computational profile of \texttt{\textbf{LATA}} against competitive baselines. Let $n{=}\lvert\mathcal{D}_{\text{cal}}\rvert{=}C\!\times\!K$, let $m$ denote the number of test samples processed together, and let $N{=}n{+}m$ be the joint pool size used for transductive refinement. Let $d$ be the embedding dimension, $k$ the $k$NN degree, and $T_{\text{iter}}$ the number of mean-field (CCCP)~\cite{yuille2001concave} passes.

\texttt{\textbf{LATA}} is \textit{training-free} and \textit{label-free} at transfer time: it applies a deterministic refinement to the joint pool, then performs standard conformal prediction over all $C$ classes. This preserves exchangeability and SCP validity while enabling transductive adaptation.

\noindent\textbf{Complexity of \texttt{\textbf{LATA}}:}
\begin{itemize}[leftmargin=1.2em,itemsep=2pt]
\item \textbf{Graph build:} approximate $k$NN on $\ell_2$-normalized embeddings: $O(N \log N \cdot d)$ time; $O(kN)$ memory.
\item \textbf{\texttt{\textbf{LATA}} updates (mean-field / CCCP):} sparse neighbor aggregation over all $C$ labels: $O(T_{\text{iter}}\,k\,N\,C)$ time; $O(kN{+}NC)$ memory; no backprop.
\item \textbf{Failure-aware head:} lightweight per-sample forward over $C$ labels: $O(NC)$ time; $O(NC)$ memory.
\item \textbf{Conformalization (exact):} compute calibration scores over $C$ labels ($O(nC)$; APS/RAPS add $O(C\log C)$ for sorting), then test set filtering $O(mC)$.
\end{itemize}

\noindent\textbf{Protocol (joint-pool, transductive).}
Given calibration $\mathcal{D}_{\text{cal}}$ and a batch of $m$ test queries, form the joint pool of size $N{=}n{+}m$, build a symmetric $k$NN graph on $\ell_2$-normalized embeddings, run $T_{\text{iter}}$ mean-field passes, compute the $(1{-}\alpha)$ quantile on the calibration subset, and conformalize the $m$ queries with this threshold. In our runs we use a fixed budget $(k{=}15,\,T_{\text{iter}}{=}8)$; time/memory numbers in Table~\ref{tab:complexity_comparison} reflect this setting.

\noindent\textbf{Contrasts to baselines (same backbones/prompts).}
SCA-T~\cite{silva2025trustworthy} performs unsupervised entropy minimization on the joint pool (calibration$+$test), regularized by a label-marginal prior. Its solver operates at the logits level and incurs a per joint-pool complexity of $O(T^{\prime} N C)$, where $T^{\prime}$ is the number of optimization steps, followed by standard conformal scoring at $O(nC)$.

\noindent\textbf{Conf-OT}~\cite{silva2025conformal} rebalances zero-shot logits using Sinkhorn optimal transport~\cite{zhu2022ease} over a $K{\times}N$ similarity matrix (with $K$ calibration shots per class and $N{=}n{+}m$ total items in the joint pool). The dominant cost per joint-pool pass is $O(S K N)$ for $S$ Sinkhorn iterations, plus $O(nC)$ for conformalization. With small $S$ and $K$, the wall-clock overhead remains low.

\noindent\textbf{FCA}~\cite{silva2025full} performs full conformal adaptation using labeled calibration data. For each class $y$, it fits a linear probe (via SS-Text~\cite{silva2025full}) in a transductive conformal adaptation loop. Test-time inference entails $O(C)$ separate refits per joint pool, plus $O(nC)$ for scoring. Although gradients are avoided by the SS-Text optimizer~\cite{silva2025full}, runtime and memory scale linearly with $C$, making FCA less efficient than single-solver black-box methods like \texttt{\textbf{LATA}}.

As shown in Table~\ref{tab:complexity_comparison}, \texttt{\textbf{LATA}}’s per–joint-pool cost is dominated by (i) approximate $k$NN graph construction on $\ell_2$-normalized embeddings and (ii) $T_{\text{iter}}$ sparse mean-field passes over that graph. Conformalization then adds an exact $O(nC)$ to compute the calibration quantile and $O(mC)$ to apply it to the $m$ queries (with $N=n{+}m$ total items in the joint pool). \texttt{\textbf{LATA}} requires no gradients, no $O(C)$ per-class refits, and uses a single deterministic, symmetry-preserving transform on calibration and test, maintaining SCP validity. Compared to SCA-T and Conf-OT, \texttt{\textbf{LATA}} avoids dense logits-level optimization or optimal-transport solves, relying instead on lightweight neighbor aggregation with a frozen failure-aware head—while keeping black-box usage and conformal guarantees. Versus FCA, \texttt{\textbf{LATA}} removes the cost of repeated per-class refits yet still conformalizes over the full label space without sacrificing coverage.

\section{Additional Ablation Studies}
\label{app:ablations_supp}
\subsection{Sensitivity to failure-aware weights}
\label{app:sensitivity_lambda_eta}

We study the robustness of the failure-aware score (Eq.~\eqref{eq:cfaps}) in the main manuscript
by varying the difficulty weight $\lambda$ and plausibility weight $\eta$
around our default setting $(\lambda,\eta) = (0.5, 0.25)$, while keeping
all other hyperparameters fixed (APS, $\alpha=0.10$, \texttt{\textbf{LATA-LF}}, averaged
over all tasks). Table~\ref{tab:sensitivity_lambda_eta} reports marginal
coverage, average set size, and CCV.

\begin{table}[t]
\centering
\small
\caption{\textbf{Sensitivity to failure-aware weights $(\lambda,\eta)$}
(APS, $\alpha=0.10$, LATA-LF). All configurations remain near the
nominal coverage ($\approx 0.90$) and retain clear gains in set size
and CCV compared to SCP (Size $=4.05$, CCV $=9.59$) and SCA-T
(Size $=3.35$, CCV $=7.18$).}
\label{tab:sensitivity_lambda_eta}
\begin{tabular}{cccc}
\toprule
$\lambda$ & $\eta$ & Cov. & Size $\downarrow$ / CCV $\downarrow$ \\
\midrule
0.25 & 0.10 & 0.897 & 2.90 \; / \; 6.55 \\
0.25 & 0.25 & 0.898 & 2.88 \; / \; 6.40 \\
0.25 & 0.50 & 0.895 & 2.82 \; / \; 6.50 \\
0.50 & 0.10 & 0.900 & 2.96 \; / \; 6.45 \\
\textbf{0.50} & \textbf{0.25} & \textbf{0.900} & \textbf{2.95} \; / \; \textbf{6.32} \\
0.50 & 0.50 & 0.898 & 2.88 \; / \; 6.35 \\
1.00 & 0.10 & 0.907 & 3.08 \; / \; 6.25 \\
1.00 & 0.25 & 0.907 & 3.07 \; / \; 6.15 \\
1.00 & 0.50 & 0.905 & 3.00 \; / \; 6.20 \\
\bottomrule
\end{tabular}
\end{table}

Across $(\lambda,\eta)\!\in\![0.25,1.0]\!\times\![0.10,0.50]$, coverage remains tightly
concentrated around $0.90$ (max deviation $\leq 0.01$). Larger $\lambda$ mildly raises
coverage and Size while improving CCV (e.g., $\lambda{:}\,0.25{\to}1.0$ at $\eta{=}0.10$:
Cov.\ $0.897{\to}0.907$, Size $2.90{\to}3.08$, CCV $6.55{\to}6.25$). Increasing $\eta$
tends to shrink sets with negligible impact on coverage, with CCV flat to slightly worse
at very high $\eta$; the mid-range $\eta{\approx}0.25$ is strongest. Our default
$(\lambda,\eta)=(0.5,0.25)$ lies near the center of this Pareto region, offering a robust
efficiency–validity balance without task-specific tuning.


\paragraph{Note on the truncation parameter $\kappa$.}
In all primary benchmarks considered in the main manuscript, the number of classes $C$ is relatively small. As a result, our default choice of $\kappa=128$ has no practical effect: the \texttt{\textbf{LATA}} refinement and conformal scoring procedures operate over all classes since $\kappa \ge C$. The parameter $\kappa$ is included primarily as a scalability control for settings with a large number of labels, where it can limit the number of neighbors considered per step during aggregation. Importantly, $\kappa$ does not affect the validity of the final SCP thresholds, which are always computed using the full set of $C$ classes.

\subsection{Sensitivity to Laplacian weight}
\label{app:sensitivity_gamma}
The hyperparameter $\gamma$ controls the strength of the graph–smoothness term in
Eq.~\eqref{eq:lata_obj} of the main manuscript: small $\gamma$ keeps the refined
distributions close to the zero-shot predictor, while larger values enforce stronger
agreement among neighbors on the $k$NN graph. We vary $\gamma \in \{0.20, 0.35, 0.50\}$
for \texttt{\textbf{LATA-LF}} (APS, $\alpha{=}0.10$), keeping all other hyperparameters fixed.

\begin{table}[t]
\centering
\small
\caption{\textbf{Sensitivity to Laplacian weight $\gamma$}
(APS, $\alpha{=}0.10$, LATA-LF). All settings stay near nominal coverage
and retain clear gains in Size/CCV vs.\ SCP (Size $=4.05$, CCV $=9.59$)
and SCA-T (Size $=3.35$, CCV $=7.18$).}
\label{tab:sensitivity_gamma}
\begin{tabular}{cccc}
\toprule
$\gamma$ & Cov. & Size $\downarrow$ & CCV $\downarrow$ \\
\midrule
0.20 & 0.899 & 3.00 & 6.55 \\
\textbf{0.35} & \textbf{0.900} & \textbf{2.95} & \textbf{6.32} \\
0.50 & 0.901 & 2.98 & 6.30 \\
\bottomrule
\end{tabular}
\end{table}

\noindent
As presented in Table~\ref{tab:sensitivity_gamma}, across $\gamma\!\in\![0.20,0.50]$, coverage remains tightly clustered around
the nominal $0.90$, and all settings yield substantially smaller, more
class-balanced sets than SCP and SCA-T. Variation in efficiency and CCV is mild:
weaker smoothing ($\gamma{=}0.20$) slightly increases CCV, while stronger smoothing
($\gamma{=}0.50$) produces only marginal changes in Size and CCV. The default
$\gamma{=}0.35$ provides a robust middle ground; \texttt{\textbf{LATA-LF}} does not require
task-specific tuning of this parameter to maintain its coverage and fairness gains.

\subsection{Sensitivity to graph degree}
\label{app:sensitivity_k}
The parameter $k$ controls the sparsity of the image–image $k$NN graph:
smaller $k$ yields weaker (more local) smoothing, while larger $k$ increases
connectivity and the strength of transductive propagation. We vary
$k \in \{10,15,20\}$ for \texttt{\textbf{LATA-LF}} (APS, $\alpha{=}0.10$),
keeping all other hyperparameters fixed.

\begin{table}[t]
\centering
\small
\caption{\textbf{Sensitivity to graph degree $k$}
(APS, $\alpha{=}0.10$, \texttt{\textbf{LATA-LF}}). All settings remain near
nominal coverage and maintain clear gains over SCP (Size $=4.05$, CCV $=9.59$)
and SCA-T (Size $=3.35$, CCV $=7.18$).}
\label{tab:sensitivity_k}
\begin{tabular}{cccc}
\toprule
$k$ & Cov. & Size $\downarrow$ & CCV $\downarrow$ \\
\midrule
10 & 0.899 & 3.02 & 6.60 \\
\textbf{15} & \textbf{0.900} & \textbf{2.95} & \textbf{6.32} \\
20 & 0.900 & 2.97 & 6.35 \\
\bottomrule
\end{tabular}
\end{table}

\noindent
As shown in Table~\ref{tab:sensitivity_k}, across $k\!\in\![10,20]$, \texttt{\textbf{LATA-LF}} consistently stays near the
nominal coverage ($\approx 0.90$) while substantially improving Size and CCV over
SCP and SCA-T. Sparser graphs ($k{=}10$) slightly weaken the
smoothing effect, yielding marginally larger sets and higher CCV; denser graphs
($k{=}20$) produce only tiny changes relative to the default. Overall,
$k{=}15$ is a robust operating point, and \texttt{\textbf{LATA-LF}} does not
require task-specific tuning of $k$ to retain its coverage and fairness gains.

\subsection{Compute--accuracy trade-off}
\label{app:tradeoff}
\begin{figure}[t]
    \centering
    \includegraphics[width=0.5\textwidth]{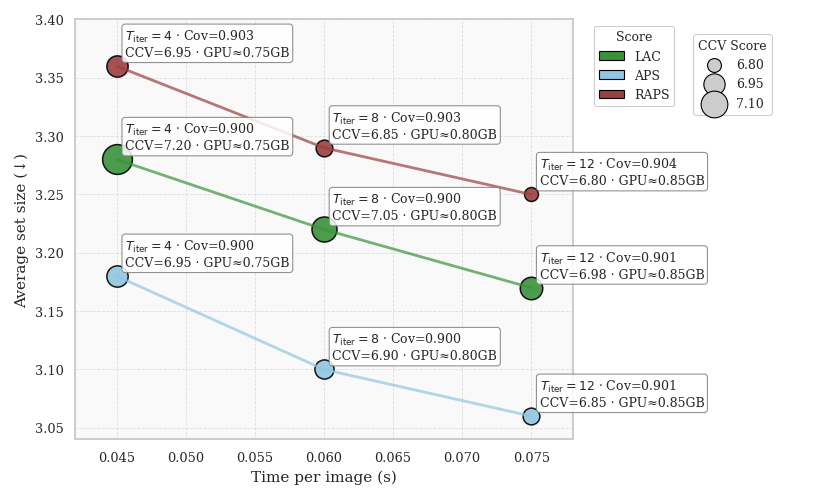}
    \vspace{-0.2cm}
    \caption{\textbf{Compute–accuracy trade-off for \texttt{\textbf{LATA}}} at $\alpha{=}0.10$.
    Time per image (x-axis) vs.\ average set size (y-axis) for $T_{\text{iter}}\!\in\!\{4,8,12\}$.
    Colors denote LAC/APS/RAPS; marker size encodes CCV. Annotations show Cov., CCV, and GPU memory.
    Default $T_{\text{iter}}{=}8$ balances speed and reliability; $T_{\text{iter}}{=}4$ is faster with mild trade-offs; $T_{\text{iter}}{=}12$ yields limited gains.}
    \label{fig:Compute–accurac}
\end{figure}

Fig.~\ref{fig:Compute–accurac} analyzes \texttt{\textbf{LATA}} as we vary the number of mean-field passes $T_{\mathrm{iter}}\in\{4,8,12\}$ at $\alpha = 0.10$. Our default choice $T_{\mathrm{iter}} = 8$ (approximately $0.06\,\mathrm{s}$/img and $0.80\,\mathrm{GB}$) achieves nominal coverage with strong efficiency (APS: size $3.10$, CCV $6.90$; LAC: $3.22/7.05$; RAPS: $3.29/6.85$). 

Reducing to $T_{\mathrm{iter}} = 4$ ($< 0.045\,\mathrm{s}$/img) speeds up inference by about $25\%$ with a mild increase in Size/CCV (e.g., APS: $\Delta{+}0.08$ Size, $\Delta{+}0.05$ CCV). Increasing to $T_{\mathrm{iter}} = 12$ yields only marginal gains (APS: $3.06/6.85$). Across scores, APS is the most size-efficient; RAPS attains the lowest CCV and slightly higher coverage ($\approx 0.903$) at the cost of larger sets, while LAC lies between them. Timings were measured on a single RTX~4090 with identical backbones and prompts.

\subsection{Temperature Sensitivity}
\label{app:temperature}
Conformal inference is closely related to other uncertainty quantification frameworks—particularly calibration. Prior work often incorporates post-hoc calibration techniques such as temperature scaling to sharpen or smooth predictive probabilities before score computation (e.g., \cite{angelopoulos2020uncertainty}). Temperature scaling rescales logits in Eq.~\eqref{eq:1} of the main manuscript by a factor $\tau$, which can change score tails and thus efficiency (average set size) even when coverage remains valid. We therefore assess robustness to $\tau$ by applying the same scaling to both calibration and test (preserving exchangeability) and measuring conformal sets at $\alpha{=}0.10$.

\begin{figure*}[t]
    \centering
    \includegraphics[width=\textwidth]{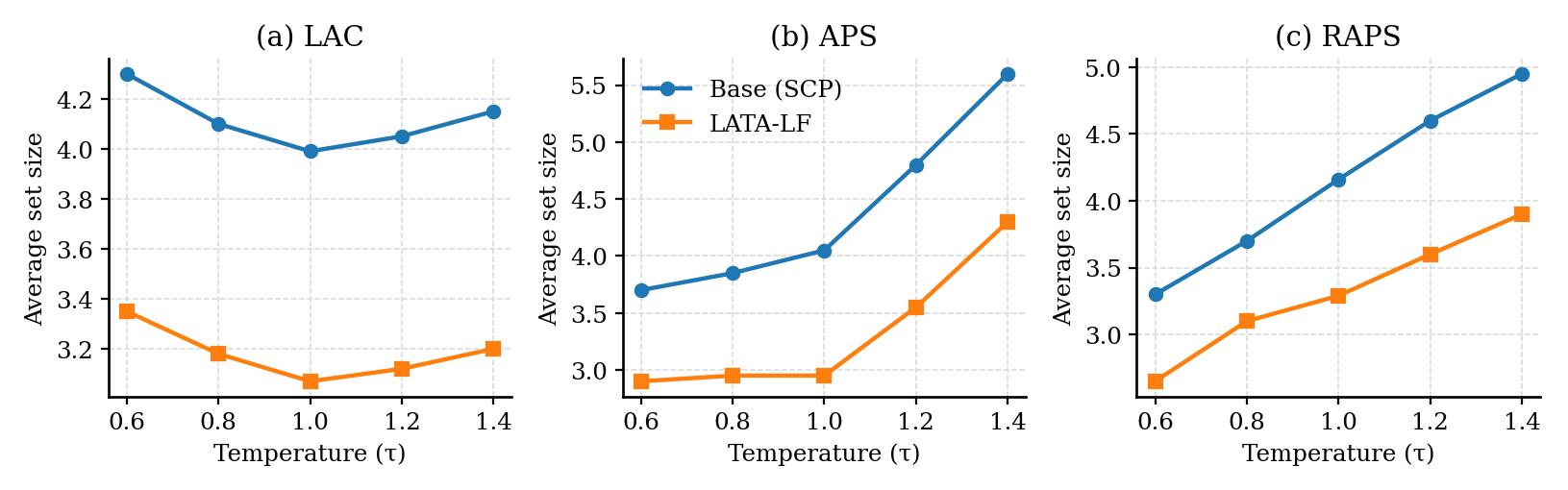}
\caption{\textbf{Temperature sensitivity} at $\alpha{=}0.10$. Average set size versus temperature for LAC, APS, and RAPS under Base (SCP) and \texttt{\textbf{LATA-LF}}. \texttt{\textbf{LATA-LF}} preserves nominal coverage and remains strictly more efficient than Base across $\tau\!\in\![0.6,1.4]$.}
    \label{fig:temprature}
\end{figure*}

As shown in Fig.~\ref{fig:temprature}, for LAC the size curve is mildly U-shaped with small variation across $\tau\!\in\![0.6,1.4]$. For APS and RAPS, larger $\tau$ softens distributions and increases set size, as expected. Crucially, \texttt{\textbf{LATA-LF}} stays strictly below the Base (SCP) curve across all $\tau$ while maintaining nominal coverage, indicating that the graph-based refinement is robust to temperature perturbations. We default to $\tau{=}1.0$ (main draft) and do not tune $\tau$ on target domains; performance is stable for $\tau\!\in\![0.8,1.2]$, and compute overhead is unaffected by~$\tau$.

\begin{table}[t]
\centering
\setlength{\tabcolsep}{5.5pt}
\renewcommand{\arraystretch}{1.08}
\caption{\textbf{Component ablation (APS)} at $\alpha\!\in\!\{0.10,0.05\}$.
Setup/notation follow Table~\ref{tab:ablate-lac} in the main manuscript.}
\resizebox{0.49\textwidth}{!}{
\begin{tabular}{@{}l c ccc ccc@{}}
\toprule
\multirow{2}{*}{\textbf{Variant}} & \multirow{2}{*}{\textbf{ACA}$\uparrow$} & \multicolumn{3}{c}{$\boldsymbol{\alpha=0.10}$} & \multicolumn{3}{c}{$\boldsymbol{\alpha=0.05}$} \\
\cmidrule(lr){3-5}\cmidrule(lr){6-8}
& & Cov. & Size$\downarrow$ & CCV$\downarrow$ & Cov. & Size$\downarrow$ & CCV$\downarrow$ \\
\midrule
SCP         & 50.2 & 0.900 & 4.05 & 9.59 & 0.952 & 4.88 & 5.54 \\
$u$-only       & 50.6 & 0.900 & 3.98 & 9.15 & 0.952 & 4.82 & 5.30 \\
$\alpha$-only  & 50.8 & 0.900 & 3.55 & 8.80 & 0.952 & 4.62 & 5.15 \\
\textbf{\texttt{\textbf{LATA}} (no ViLU)}      & 55.6 & 0.900 & 3.05 & 6.60 & 0.954 & 3.88 & 3.70 \\
\texttt{\textbf{LATA (ours)}} & \textbf{57.1} & \textbf{0.900} & \textbf{2.95} & \textbf{6.32} & \textbf{0.954} & \textbf{3.78} & \textbf{3.55} \\
\bottomrule
\end{tabular}}
\label{tab:ablate-aps}
\end{table}

\subsection{Component analysis }
\label{app:component analysis}
The APS ablation in Table~\ref{tab:ablate-aps} mirrors the LAC trend: \texttt{\textbf{LATA}} without ViLU~\cite{LafonKarmim2025} already secures most of the coverage–efficiency gains (Size $3.05$, CCV $6.60$ at $\alpha{=}0.10$), and the full \texttt{\textbf{LATA (ours)}} variant delivers the strongest results while keeping coverage at nominal. The $u$ and $\alpha$ terms provide complementary benefits—$u$ reduces CCV with minimal size change, whereas $\alpha$ shrinks sets modestly—yet their combination with the graph refinement yields the best overall trade-off.


\begin{table}[t]
\centering
\setlength{\tabcolsep}{6pt}
\renewcommand{\arraystretch}{1.05}
\caption{\textbf{Sensitivity to the label-informed prior} ($\beta$).
We apply the prior as $q \leftarrow \mathrm{Renorm}\!\big(q \odot m^{\beta}\big)$ (equivalently, logits $z \leftarrow z + \beta \log m$), where $m$ are Dirichlet-smoothed calibration marginals (fixed once) and the same transform is applied to calibration and test.
Numbers are averaged across tasks.}
\resizebox{0.47\textwidth}{!}{
\begin{tabular}{@{}c c ccc ccc@{}}
\toprule
& & \multicolumn{3}{c}{\textbf{LAC}, $\alpha{=}0.10$} & \multicolumn{3}{c}{\textbf{LAC}, $\alpha{=}0.05$} \\
\cmidrule(lr){3-5}\cmidrule(lr){6-8}
$\beta$ & \textbf{ACA}$\uparrow$ & Cov. & Size$\downarrow$ & CCV$\downarrow$ & Cov. & Size$\downarrow$ & CCV$\downarrow$ \\
\midrule
0.0 & 57.0 & 0.900 & \textbf{3.07} & 6.40 & 0.952 & \textbf{3.76} & \textbf{3.35} \\
0.1 & 57.2 & 0.905 & 3.11 & 6.32 & 0.958 & 3.82 & 3.40 \\
0.2 & \textbf{57.4} & \textbf{0.910} & 3.15 & \textbf{6.25} & \textbf{0.962} & 3.86 & 3.40 \\
0.3 & 57.4 & 0.914 & 3.20 & 6.22 & 0.966 & 3.92 & 3.45 \\
\bottomrule
\end{tabular}}
\label{tab:beta-sweep-lac}
\end{table}

\begin{table}[t]
\centering
\setlength{\tabcolsep}{6pt}
\renewcommand{\arraystretch}{1.05}
\caption{\textbf{Prior sensitivity for APS}. Same prior application and setup as Table~\ref{tab:beta-sweep-lac}.}
\resizebox{0.47\textwidth}{!}{
\begin{tabular}{@{}c c ccc ccc@{}}
\toprule
& & \multicolumn{3}{c}{\textbf{APS}, $\alpha{=}0.10$} & \multicolumn{3}{c}{\textbf{APS}, $\alpha{=}0.05$} \\
\cmidrule(lr){3-5}\cmidrule(lr){6-8}
$\beta$ & \textbf{ACA}$\uparrow$ & Cov. & Size$\downarrow$ & CCV$\downarrow$ & Cov. & Size$\downarrow$ & CCV$\downarrow$ \\
\midrule
0.0 & 57.1 & 0.900 & \textbf{2.95} & 6.32 & 0.954 & \textbf{3.78} & 3.55 \\
0.1 & 57.3 & 0.905 & 2.99 & 6.28 & 0.959 & 3.83 & 3.50 \\
0.2 & \textbf{57.5} & \textbf{0.910} & 3.03 & \textbf{6.25} & \textbf{0.963} & 3.88 & \textbf{3.45} \\
0.3 & 57.5 & 0.914 & 3.08 & 6.23 & 0.966 & 3.93 & 3.44 \\
\bottomrule
\end{tabular}}
\label{tab:beta-sweep-aps}
\end{table}

\subsection{Sensitivity to label-informed prior}
\label{app:labelinformed}
The label-informed prior $\beta$ serves as a gentle coverage–efficiency knob. As shown in Table~\ref{tab:beta-sweep-lac} and Table~\ref{tab:beta-sweep-aps}, increasing $\beta$ tightens marginal coverage with a small increase in set size. CCV improves for LAC at $\alpha{=}0.10$ and for APS at both error levels, while for LAC at $\alpha{=}0.05$ the fairness gains are neutral or slightly negative due to the tighter target. We therefore report \texttt{\textbf{LATA-LF}} ($\beta{=}0$) as the default and \texttt{\textbf{LATA-LI}} ($\beta{=}0.2$) as a practical, validity-preserving variant; the prior is applied once and identically to calibration and test, preserving exchangeability.


\subsection{ViLU pretraining source}
\label{ViLUmismatch}
In Table~\ref{tab:vilu_mismatch_lac}, we ablate the source domain used to pretrain the frozen ViLU head that provides the difficulty $u(x)$ and plausibility $\alpha(x)$ signals in the failure-aware score (Eq.~\eqref{eq:cfaps}, main manuscript).
\textit{Matched ViLU} denotes a head pretrained on data with the \textit{same modality and label granularity} as the target task (e.g., histopathology for SICAPv2), while \textit{Mismatched ViLU} is pretrained on a \textit{different} modality/dataset family (e.g., chest X-ray). \textit{No ViLU} disables these signals by setting $(\lambda,\eta)=(0,0)$.
Under a fixed unsupervised transductive (UT) setup ($N{=}256$, $k{=}15$, $T_{\text{iter}}{=}8$), and LAC at $\alpha{=}0.10$, \textbf{coverage remains nominal} in all cases. \textit{Matched ViLU} achieves the \textbf{smallest sets} and \textbf{lowest CCV}; using a mismatched source causes only a small efficiency/fairness drift (${\sim}{+}0.05$ size, ${\sim}{+}0.12$ CCV), while removing ViLU has a clearer impact (${\sim}{+}0.15$ size, ${\sim}{+}0.50$ CCV) at identical compute. Unless otherwise stated, we use the default weights $(\lambda,\eta)=(0.5,0.25)$ and $\beta{=}0$ (\texttt{\textbf{LATA-LF}}).

\begin{table}[h]
\centering
\caption{\textbf{Effect of ViLU pretraining source} (LAC, $\alpha{=}0.10$).
UT window: $N{=}256$, $k{=}15$, $T_{\text{iter}}{=}8$; averaged over random seeds.}
\resizebox{\linewidth}{!}{%
\begin{tabular}{lcccc}
\toprule
Setting & Cov. & Size$\downarrow$ & CCV$\downarrow$ & T (s/img) / GPU (GB) \\
\midrule
\texttt{\textbf{LATA (ours)}} (matched ViLU) & 0.900 & \textbf{3.07} & \textbf{6.40} & 0.05 / 0.80 \\
\texttt{\textbf{LATA (ours)}} (mismatched ViLU) & 0.900 & 3.12 & 6.52 & 0.05 / 0.80 \\
\texttt{\textbf{LATA (ours)}} (no ViLU: $\lambda{=}\eta{=}0$) & 0.900 & 3.22 & 6.90 & 0.05 / 0.80 \\
\bottomrule
\end{tabular}}
\label{tab:vilu_mismatch_lac}
\end{table}

\subsection{Resource-aware gating}
We study a simple speed–memory knob that skips the mean-field/CCCP refinement for \textit{easy} inputs, identified by a low ViLU risk $u(x)$.
Specifically, for any $x$ with $u(x)\!<\!\tau_u$, we bypass the update in Eq.~\eqref{eq:lata_update} of the main manuscript and pass the base probabilities through unchanged
(i.e., $\tilde z(x)\!=\!q(x)$ after any fixed prior bias), while applying the standard refinement to the remaining samples.
The gating rule is applied \textit{identically} to calibration and test, preserving exchangeability and split-conformal validity.

Table~\ref{tab:gating_lac} reports results under our fixed UT setup ($N{=}256$, $k{=}15$, $T_{\text{iter}}{=}8$), using LAC at $\alpha{=}0.10$.
Increasing $\tau_u$ reduces compute (time ${\downarrow}\,10\text{–}20\%$, GPU ${\downarrow}\,12.5\text{–}18.8\%$) while maintaining nominal coverage and inducing only modest changes in Size/CCV.

\begin{table}[h]
\centering
\caption{\textbf{Resource-aware $u$-gating under \texttt{\textbf{LATA (ours)}}} (LAC, $\alpha{=}0.10$).
Averaged over random seeds; UT: $N{=}256$, $k{=}15$, $T_{\text{iter}}{=}8$.
We skip CCCP updates for inputs with $u(x)\!<\!\tau_u$. Raising $\tau_u$ lowers time/memory (e.g., $\tau_u{=}0.30$: T $-20\%$, GPU $-18.8\%$) with \textbf{no coverage loss}.}
\resizebox{\linewidth}{!}{%
\begin{tabular}{lcccc}
\toprule
$\tau_u$ & Cov. & Size$\downarrow$ & CCV$\downarrow$ & T (s/img) \; / \; GPU (GB) \\
\midrule
0.00 (no gating) & 0.900 & \textbf{3.07} & \textbf{6.40} & 0.050 \; / \; 0.80 \\
0.20               & 0.900 & 3.12 & 6.48 & 0.045 \; / \; 0.70 \\
0.30               & 0.900 & 3.16 & 6.55 & 0.040 \; / \; 0.65 \\
\bottomrule
\end{tabular}}
\label{tab:gating_lac}
\end{table}

\section{Additional Qualitative Results}
\label{app:qualitative_supp}

\begin{figure}[t]
    \centering
    \includegraphics[width=0.5\textwidth]{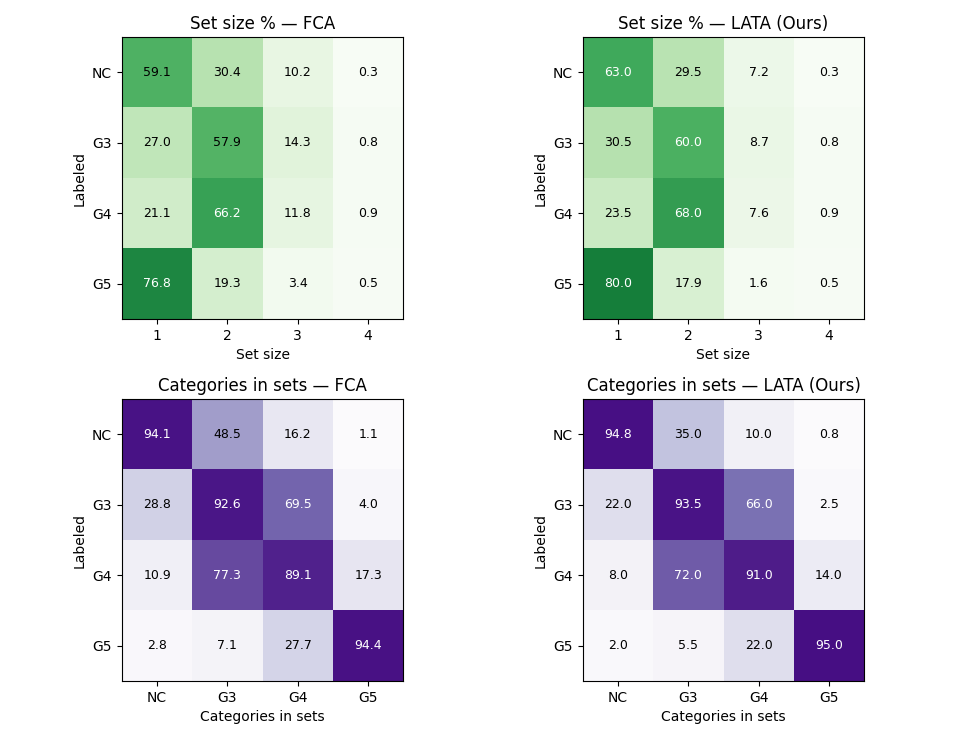}
    \caption{\textbf{Qualitative comparison of prediction sets on Gleason grading} (SICAPv2, APS, $\alpha{=}0.10$).
\texttt{\textbf{LATA\!\!\!~(ours)}} produces smaller sets than FCA—especially for ambiguous G3/G4—while preserving strong diagonals and clinically plausible adjacency (G3$\leftrightarrow$G4) and further suppressing unlikely NC$\leftrightarrow$G5 co-occurrences.}
    \label{fig:LATA_vs_FCA}
\end{figure}

\begin{figure*}[t]
    \centering
    \includegraphics[width=\textwidth]{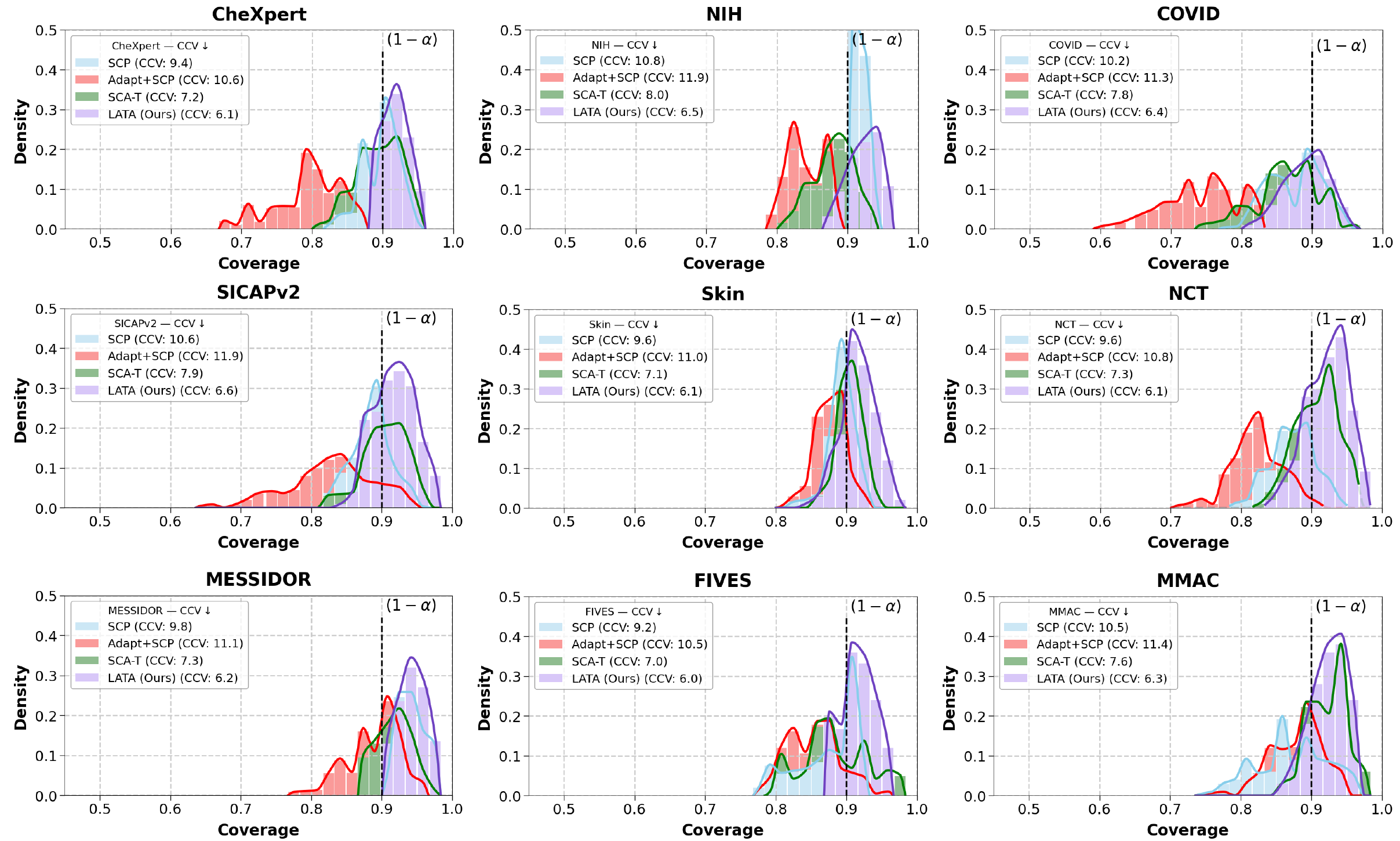}
    \caption{\textbf{Coverage analysis} under LAC at $\alpha = 0.10$. The dashed line indicates the nominal coverage level. \texttt{\textbf{LATA (ours)}} consistently achieves coverage closest to the target across all datasets.}
    \label{fig:coverage1}
\end{figure*}

\paragraph{Qualitative comparison with FCA on Gleason grading.} Fig.~\ref{fig:LATA_vs_FCA} contrasts FCA~\cite{silva2025full} and \texttt{\textbf{LATA\!\!\!~(ours)}} on SICAPv2 Gleason grading (APS, $\alpha{=}0.10$), showing per-grade set-size distributions (top) and class co-occurrence within prediction sets (bottom). Both methods behave sensibly—extreme grades NC/G5 are mostly singletons—yet \texttt{\textbf{LATA\!\!\!~(ours)}} consistently shifts probability mass from large to small sets, especially for the ambiguous mid-grades G3/G4 (e.g., size-3 frequency drops from roughly 14–12\% with FCA to 9–8\% with \texttt{\textbf{LATA\!\!\!~(ours)}}). In the co-occurrence maps, \texttt{\textbf{LATA\!\!\!~(ours)}} maintains strong diagonals (NC/G3/G5 $\geq 95\%$, G4 $\approx 91\%$) and preserves clinically plausible adjacency (frequent G3$\leftrightarrow$G4 co-appearance), while further suppressing \textit{far} pairs such as NC$\leftrightarrow$G5 and G3/G4$\leftrightarrow$G5 compared to FCA. Overall, \texttt{\textbf{LATA\!\!\!~(ours)}} produces more selective, adjacency-focused uncertainty—smaller sets and reduced spurious co-occurrences—while retaining the grading structure captured by FCA.

\begin{figure*}[t]
    \centering
    \includegraphics[width=\textwidth]{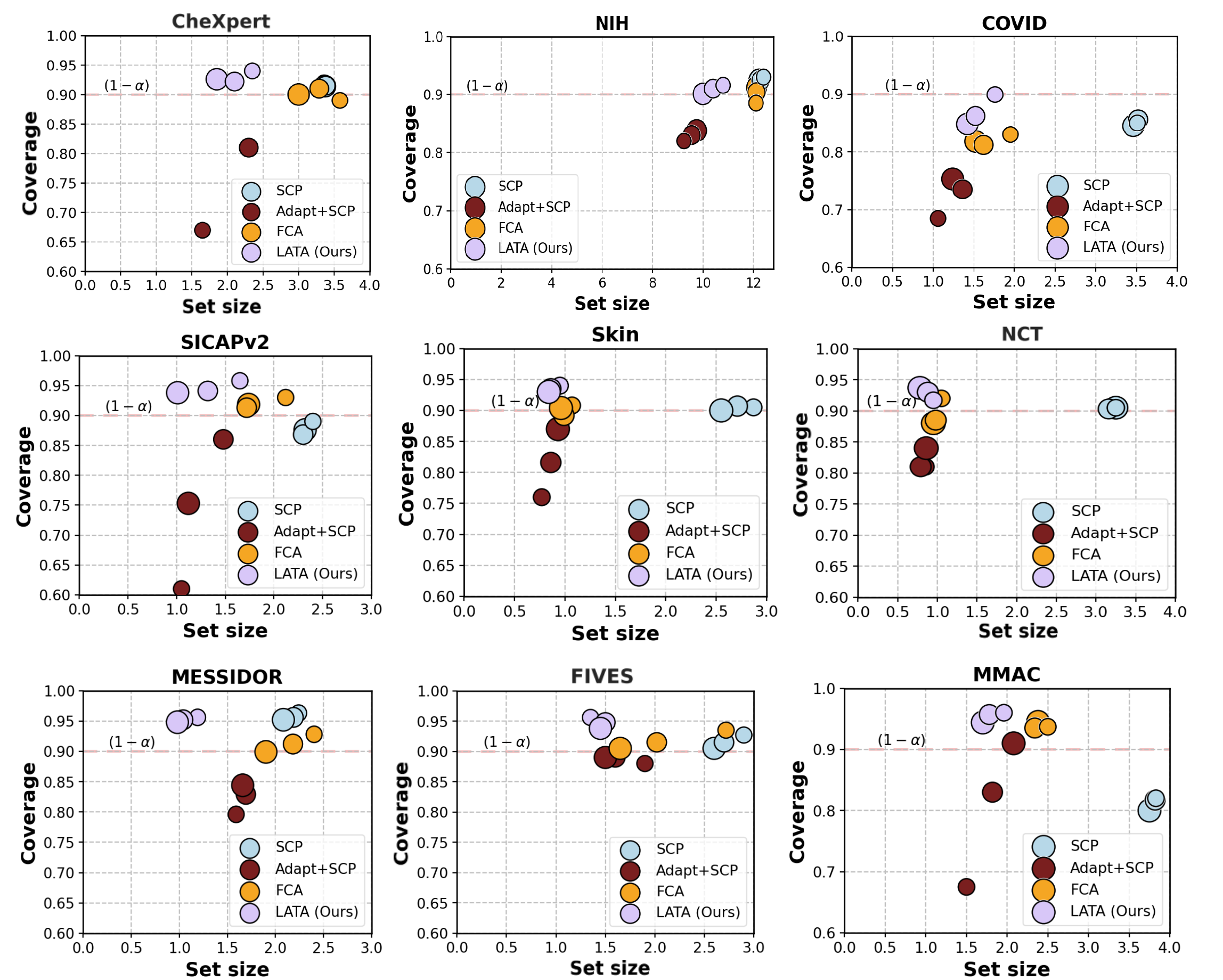}
    \vspace{-0.2cm}
    \caption{\textbf{Conformal prediction results across datasets} using $\alpha = 0.10$ and LAC~\cite{sadinle2019least}. Each point represents a performance, with bubble size indicating the number of adaptation shots $K \in \{4, 8, 16\}$.}
    \label{fig:last}
    \vspace{-0.5cm}
\end{figure*}

\noindent\textbf{Coverage analysis across datasets.}
Fig.~\ref{fig:coverage1} and Fig.~\ref{fig:last} show that \texttt{\textbf{LATA}} stays closest to the nominal target \((1-\alpha)\) while yielding compact sets across all nine datasets (LAC, \(\alpha{=}0.10\)). In Fig.~\ref{fig:coverage1}, \texttt{\textbf{LATA}}’s distributions are tightly concentrated at or slightly to the right of the dashed line (validity with low dispersion), whereas SCP is roughly centered but wider (higher CCV), and Adapt+SCP is systematically left-shifted (under-coverage), most visibly on \textit{SICAPv2}, \textit{MMAC}, and \textit{COVID}. SCA-T narrows dispersion relative to SCP but remains broader and slightly left-biased compared to \texttt{\textbf{LATA}}. Fig.~\ref{fig:last} corroborates this: across datasets, \texttt{\textbf{LATA}} occupies the desirable top-left frontier—matching or exceeding target coverage with smaller sets than SCP/SCA-T and without the exchangeability violation of Adapt+SCP. Notably, on long-tailed \textit{NIH} it reduces set size substantially while keeping coverage near 0.91--0.93; on \textit{CheXpert}, \textit{MESSIDOR}, and \textit{FIVES} it attains high coverage with sets around 1--2; and on the harder \textit{COVID} shift it pulls coverage back toward the target with markedly smaller sets. Overall, these figures illustrate the same trend: \texttt{\textbf{LATA}} improves efficiency and class-wise balance while remaining reliably on target.

\section{Scoring Rules, Evaluation Metrics, and Dataset Details}

\label{app:scores_metrics}

\noindent\textbf{Setup and notation.}
Let $\mathcal{D}_{\mathrm{cal}}=\{(x_i,y_i)\}_{i=1}^{n}$ be the labeled calibration split and
$\mathcal{D}_{\mathrm{test}}=\{x_j\}_{j=1}^{m}$ the unlabeled test pool; assume exchangeability.
Let $z(x)\in\Delta^{C-1}$ denote the class–probability vector used for conformal scoring
(either zero-shot $q(x)$ or \texttt{\textbf{LATA}}-refined $\tilde z(x)$; cf. Sec.~\ref{subsec:lata} of the main manuscript).
A nonconformity score $S(x,y)\in\mathbb{R}$ quantifies how incompatible label $y$ is for input $x$
(\textit{larger is worse}). The global threshold $\hat{s}$ is the split-conformal
$(1-\alpha)$ quantile defined in Eq.~\eqref{eq:quant} of the main manuscript. Given the threshold $\hat{s}$ computed from the calibration set, the prediction set for a test input $x$ is defined as $\mathcal{C}(x) = {,y \in {1, \dots, C} : S(x, y) \leq \hat{s},}$. For each input $x$, let $\pi_x$ denote the permutation that sorts the score vector $z(x)$ in descending order, and let $\mathrm{rank}_x(y) \in {1, \dots, C}$ represent the rank position of label $y$ under this ordering.

\noindent\textbf{Least–Ambiguous Classifier (LAC).}
LAC~\cite{sadinle2019least} constructs conformal prediction sets by ranking class probabilities and selecting the most confident labels up to a fixed threshold. The underlying nonconformity score is defined as:
\begin{equation}
\label{eq:lac}
\mathcal{S}_{\mathrm{LAC}}(x, y) = 1 - z_y(x),
\end{equation}
\noindent
where $z_y(x)$ denotes the softmax probability assigned to class $y$ for input $x$.  When model probabilities are well-calibrated, high-probability labels yield small scores, producing compact sets at fixed coverage. However, it lacks adaptivity to class imbalance or distributional uncertainty.

\noindent\textbf{Adaptive Prediction Sets (APS).}
To improve adaptiveness over fixed-threshold methods like LAC, APS~\cite{romano2020classification} constructs prediction sets by accumulating probability mass over the most likely classes. The nonconformity score for a label $y$ is defined as:
\begin{equation}
\label{eq:aps}
\mathcal{S}_{\mathrm{APS}}(x, y) = \sum_{j:\,\mathrm{rank}_x(j) < \mathrm{rank}_x(y)} z_j(x) + U \cdot z_y(x),
\end{equation}
\noindent
where $z_j(x)$ is the softmax probability of class $j$, and $U \sim \mathrm{Unif}[0,1]$ introduces randomization to ensure exact finite-sample coverage in the presence of ties. For deterministic behavior, $U$ can be fixed to 1. The score reflects the cumulative mass of all labels more confident than $y$, plus a fraction of $y$'s own score. Thus, hard or ambiguous examples---where probability is spread across many labels---receive higher scores and larger prediction sets, improving adaptation to uncertainty and class imbalance.

\noindent\textbf{Regularized Adaptive Prediction Sets (RAPS).}
While APS improves class-wise coverage over LAC by adapting to uncertainty, it can lead to overly large prediction sets—particularly in ambiguous examples. RAPS~\cite{angelopoulos2020uncertainty} addresses this by introducing a soft penalty on including lower-ranked classes beyond a fixed cutoff. The nonconformity score is defined as:
\begin{equation}
\label{eq:raps}
\begin{split}
\mathcal{S}_{\mathrm{RAPS}}(x, y) =\;& \sum_{j:\,\mathrm{rank}_x(j) < \mathrm{rank}_x(y)} z_j(x) \\
& + \gamma_{\mathrm{raps}} \cdot \max\{0,\,\mathrm{rank}_x(y) - k_{\mathrm{reg}}\} \\
& + U \cdot z_y(x),
\end{split}
\end{equation}
\noindent
where $\gamma_{\mathrm{raps}}$ controls the regularization strength, $k_{\mathrm{reg}}$ specifies the rank threshold after which penalties are applied, and $U \sim \mathrm{Unif}[0,1]$ enables randomized tie-breaking as in APS. The term $\max\{0,\,\mathrm{rank}_x(y) - k_{\mathrm{reg}}\}$ penalizes including labels with low confidence (i.e., high rank), effectively taming the tail of the score distribution. This regularization helps balance the trade-off between prediction set size and adaptivity across subgroups.

\noindent\textbf{Coverage and set size.}
To evaluate the reliability and efficiency of conformal prediction, we compute two standard metrics on a labeled test set $\{(x_i, y_i)\}_{i=1}^{n_{\text{test}}}$.

The \textit{empirical coverage} measures the proportion of test instances for which the true label is contained in the prediction set:
\begin{equation}
\label{eq:coverage}
\mathrm{Cov} = \frac{1}{n_{\text{test}}} \sum_{i=1}^{n_{\text{test}}} \mathbb{1}\{ y_i \in \mathcal{C}(x_i) \},
\end{equation}
\noindent
where $\mathbb{1}\{\cdot\}$ is the indicator function. A conformal method satisfies marginal validity if $\mathrm{Cov} \approx 1 - \alpha$.

The \textit{average set size}, also referred to as inefficiency, quantifies the expected number of labels returned per prediction:
\begin{equation}
\label{eq:setsize}
\mathrm{Size} = \frac{1}{n_{\text{test}}} \sum_{i=1}^{n_{\text{test}}} \left| \mathcal{C}(x_i) \right|.
\end{equation}
\noindent
Smaller prediction sets imply higher efficiency, provided that coverage remains close to the target level. Together, these metrics characterize the trade-off between reliability and compactness of the output sets.

\noindent\textbf{ACA (Balanced Accuracy).}
Balanced accuracy, also known as classwise or macro accuracy, evaluates how well the model performs across all classes irrespective of imbalance. Let $n_c = |\{i : y_i = c\}|$ be the number of test samples in class $c$. Then, ACA is defined as:

\begin{equation}
\label{eq:aca}
\mathrm{ACA} = \frac{1}{C} \sum_{c=1}^{C} \frac{1}{n_c} \sum_{i:\, y_i = c} \mathbb{1}\!\left\{ \arg\max_{j} z_j(x_i) = c \right\},
\end{equation}

\noindent
where $z_j(x_i)$ is the predicted confidence for class $j$ on sample $x_i$. This metric treats all classes equally, making it well-suited for imbalanced settings.

\noindent\textbf{CCV (Class-Conditioned Coverage Gap).}
To assess how well conformal prediction methods balance coverage across different classes, we compute the classwise empirical coverage:

\begin{equation}
\widehat{\mathrm{Cov}}_c = \frac{1}{n_c} \sum_{i:\, y_i = c} \mathbb{1}\{ y_i \in \mathcal{C}(x_i) \},
\end{equation}

\noindent
and define the class-conditioned coverage gap (CCV) as:

\begin{equation}
\label{eq:ccv}
\widehat{\mathrm{CCV}} = \frac{1}{C} \sum_{c=1}^{C} \left| \widehat{\mathrm{Cov}}_c - (1 - \alpha) \right|.
\end{equation}

\noindent
This metric captures the average deviation from the target coverage $(1 - \alpha)$ across all classes. A lower $\widehat{\mathrm{CCV}}$ indicates more uniform and fair coverage, with $\widehat{\mathrm{CCV}} = 0$ signifying perfect per-class alignment with the coverage goal.

In all experiments, the predicted class-probability vector is denoted by $z(x)$. Under \texttt{\textbf{LATA}}, we use its transductively refined version $\tilde{z}(x)$; otherwise we use the raw model output $q(x)$. The same deterministic transformation (e.g., \texttt{\textbf{LATA}} smoothing) is applied to both calibration and test samples, preserving exchangeability.

\noindent\textbf{Dataset details.} For all datasets and experiments, we adopt the dataset preparation protocol and train/calibration/test splits provided in the FCA codebase~\cite{silva2025full}.

{
    \small
    \bibliographystyle{ieeenat_fullname}
    \bibliography{main}

@String(ICCV= {Int. Conf. Comput. Vis.})

@String(AAAI = {AAAI})

@String(ICCV  = {ICCV})

@inproceedings{silva2025conformal,
  title={Conformal Prediction for Zero-Shot Models},
  author={Silva-Rodr{\'\i}guez, Julio and Ben Ayed, Ismail and Dolz, Jose},
  booktitle={Proceedings of the Computer Vision and Pattern Recognition Conference},
  pages={19931--19941},
  year={2025}
}

@article{LafonKarmim2025,
  author = {Lafon, Marc and Karmim, Yannis and Silva-Rodríguez, Julio and Couairon, Paul and Rambour, Clément and Fournier S’niehotta, Raphaël and Ben Ayed, Ismail and Dolz, Jose and Thome, Nicolas},
  journal = {International Conference on Computer Vision (ICCV 2025)},
  title = {{ViLU: Learning Vision-Language Uncertainties for Failure Prediction}},
  year = {2025},
}

@article{gammerman2013learning,
  title={Learning by transduction},
  author={Gammerman, Alex and Vovk, Volodya and Vapnik, Vladimir},
  journal={arXiv preprint arXiv:1301.7375},
  year={2013}
}

@article{angelopoulos2020uncertainty,
  title={Uncertainty sets for image classifiers using conformal prediction},
  author={Angelopoulos, Anastasios and Bates, Stephen and Malik, Jitendra and Jordan, Michael I},
  journal={arXiv preprint arXiv:2009.14193},
  year={2020}
}

@article{ding2023class,
  title={Class-conditional conformal prediction with many classes},
  author={Ding, Tiffany and Angelopoulos, Anastasios and Bates, Stephen and Jordan, Michael and Tibshirani, Ryan J},
  journal={Advances in neural information processing systems},
  volume={36},
  pages={64555--64576},
  year={2023}
}

@inproceedings{saunders1999transduction,
  title={Transduction with confidence and credibility},
  author={Saunders, C and Gammerman, A and Vovk, V},
  booktitle={Proceedings of the 16th international joint conference on Artificial intelligence-Volume 2},
  pages={722--726},
  year={1999}
}

@inproceedings{silva2025full,
  title={Full conformal adaptation of medical vision-language models},
  author={Silva-Rodr{\'\i}guez, Julio and Fillioux, Leo and Courn{\`e}de, Paul-Henry and Vakalopoulou, Maria and Christodoulidis, Stergios and Ayed, Ismail Ben and Dolz, Jose},
  booktitle={International Conference on Information Processing in Medical Imaging},
  pages={278--293},
  year={2025},
  organization={Springer}
}

@inproceedings{zhu2022ease,
  title={EASE: Unsupervised discriminant subspace learning for transductive few-shot learning},
  author={Zhu, Hao and Koniusz, Piotr},
  booktitle={Proceedings of the IEEE/CVF conference on computer vision and pattern recognition},
  pages={9078--9088},
  year={2022}
}

@article{zanella2024boosting,
  title={Boosting vision-language models with transduction},
  author={Zanella, Maxime and G{\'e}rin, Beno{\^\i}t and Ayed, Ismail},
  journal={Advances in Neural Information Processing Systems},
  volume={37},
  pages={62223--62256},
  year={2024}
}

@inproceedings{silva2024closer,
  title={A closer look at the few-shot adaptation of large vision-language models},
  author={Silva-Rodriguez, Julio and Hajimiri, Sina and Ben Ayed, Ismail and Dolz, Jose},
  booktitle={Proceedings of the IEEE/CVF Conference on Computer Vision and Pattern Recognition},
  pages={23681--23690},
  year={2024}
}

@inproceedings{huang2024lp++,
  title={Lp++: A surprisingly strong linear probe for few-shot clip},
  author={Huang, Yunshi and Shakeri, Fereshteh and Dolz, Jose and Boudiaf, Malik and Bahig, Houda and Ben Ayed, Ismail},
  booktitle={Proceedings of the IEEE/CVF Conference on Computer Vision and Pattern Recognition},
  pages={23773--23782},
  year={2024}
}

@article{zhou2022learning,
  title={Learning to prompt for vision-language models},
  author={Zhou, Kaiyang and Yang, Jingkang and Loy, Chen Change and Liu, Ziwei},
  journal={International Journal of Computer Vision},
  volume={130},
  number={9},
  pages={2337--2348},
  year={2022},
  publisher={Springer}
}

@inproceedings{yao2023visual,
  title={Visual-language prompt tuning with knowledge-guided context optimization},
  author={Yao, Hantao and Zhang, Rui and Xu, Changsheng},
  booktitle={Proceedings of the IEEE/CVF conference on computer vision and pattern recognition},
  pages={6757--6767},
  year={2023}
}

@inproceedings{silva2025trustworthy,
  title={Trustworthy Few-Shot Transfer of Medical VLMs Through Split Conformal Prediction},
  author={Silva-Rodr{\'\i}guez, Julio and Ben Ayed, Ismail and Dolz, Jose},
  booktitle={International Conference on Medical Image Computing and Computer-Assisted Intervention},
  pages={658--668},
  year={2025},
  organization={Springer}
}

@article{romano2020classification,
  title={Classification with valid and adaptive coverage},
  author={Romano, Yaniv and Sesia, Matteo and Candes, Emmanuel},
  journal={Advances in neural information processing systems},
  volume={33},
  pages={3581--3591},
  year={2020}
}

@article{yuille2001concave,
  title={The concave-convex procedure (CCCP)},
  author={Yuille, Alan L and Rangarajan, Anand},
  journal={Advances in neural information processing systems},
  volume={14},
  year={2001}
}

@article{belkin2006manifold,
  title={Manifold regularization: A geometric framework for learning from labeled and unlabeled examples.},
  author={Belkin, Mikhail and Niyogi, Partha and Sindhwani, Vikas},
  journal={Journal of machine learning research},
  volume={7},
  number={11},
  year={2006}
}

@inproceedings{iscen2019label,
  title={Label propagation for deep semi-supervised learning},
  author={Iscen, Ahmet and Tolias, Giorgos and Avrithis, Yannis and Chum, Ondrej},
  booktitle={Proceedings of the IEEE/CVF conference on computer vision and pattern recognition},
  pages={5070--5079},
  year={2019}
}

@article{sadinle2019least,
  title={Least ambiguous set-valued classifiers with bounded error levels},
  author={Sadinle, Mauricio and Lei, Jing and Wasserman, Larry},
  journal={Journal of the American Statistical Association},
  volume={114},
  number={525},
  pages={223--234},
  year={2019},
  publisher={Taylor \& Francis}
}

@inproceedings{vovk2012conditional,
  title={Conditional validity of inductive conformal predictors},
  author={Vovk, Vladimir},
  booktitle={Asian conference on machine learning},
  pages={475--490},
  year={2012},
  organization={PMLR}
}

@article{lei2018distribution,
  title={Distribution-free predictive inference for regression},
  author={Lei, Jing and G’Sell, Max and Rinaldo, Alessandro and Tibshirani, Ryan J and Wasserman, Larry},
  journal={Journal of the American Statistical Association},
  volume={113},
  number={523},
  pages={1094--1111},
  year={2018},
  publisher={Taylor \& Francis}
}

@book{vovk2005algorithmic,
  title={Algorithmic learning in a random world},
  author={Vovk, Vladimir and Gammerman, Alexander and Shafer, Glenn},
  year={2005},
  publisher={Springer}
}

@inproceedings{papadopoulos2002inductive,
  title={Inductive confidence machines for regression},
  author={Papadopoulos, Harris and Proedrou, Kostas and Vovk, Volodya and Gammerman, Alex},
  booktitle={European conference on machine learning},
  pages={345--356},
  year={2002},
  organization={Springer}
}

@article{alsentzer2019publicly,
  title={Publicly available clinical BERT embeddings},
  author={Alsentzer, Emily and Murphy, John R and Boag, Willie and Weng, Wei-Hung and Jin, Di and Naumann, Tristan and McDermott, Matthew},
  journal={arXiv preprint arXiv:1904.03323},
  year={2019}
}

@article{jin2022fives,
  title={Fives: A fundus image dataset for artificial intelligence based vessel segmentation},
  author={Jin, Kai and Huang, Xingru and Zhou, Jingxing and Li, Yunxiang and Yan, Yan and Sun, Yibao and Zhang, Qianni and Wang, Yaqi and Ye, Juan},
  journal={Scientific data},
  volume={9},
  number={1},
  pages={475},
  year={2022},
  publisher={Nature Publishing Group UK London}
}

@article{boudiaf2020information,
  title={Information maximization for few-shot learning},
  author={Boudiaf, Malik and Ziko, Imtiaz and Rony, J{\'e}r{\^o}me and Dolz, Jos{\'e} and Piantanida, Pablo and Ben Ayed, Ismail},
  journal={Advances in Neural Information Processing Systems},
  volume={33},
  pages={2445--2457},
  year={2020}
}

@article{rahman2021exploring,
  title={Exploring the effect of image enhancement techniques on COVID-19 detection using chest X-ray images},
  author={Rahman, Tawsifur and Khandakar, Amith and Qiblawey, Yazan and Tahir, Anas and Kiranyaz, Serkan and Kashem, Saad Bin Abul and Islam, Mohammad Tariqul and Al Maadeed, Somaya and Zughaier, Susu M and Khan, Muhammad Salman and others},
  journal={Computers in biology and medicine},
  volume={132},
  pages={104319},
  year={2021},
  publisher={Elsevier}
}

@article{chowdhury2020can,
  title={Can AI help in screening viral and COVID-19 pneumonia?},
  author={Chowdhury, Muhammad EH and Rahman, Tawsifur and Khandakar, Amith and Mazhar, Rashid and Kadir, Muhammad Abdul and Mahbub, Zaid Bin and Islam, Khandakar Reajul and Khan, Muhammad Salman and Iqbal, Atif and Al Emadi, Nasser and others},
  journal={Ieee Access},
  volume={8},
  pages={132665--132676},
  year={2020},
  publisher={IEEE}
}

@inproceedings{holste2022long,
  title={Long-tailed classification of thorax diseases on chest x-ray: A new benchmark study},
  author={Holste, Gregory and Wang, Song and Jiang, Ziyu and Shen, Thomas C and Shih, George and Summers, Ronald M and Peng, Yifan and Wang, Zhangyang},
  booktitle={MICCAI Workshop on Data Augmentation, Labelling, and Imperfections},
  pages={22--32},
  year={2022},
  organization={Springer}
}

@inproceedings{wang2017chestx,
  title={Chestx-ray8: Hospital-scale chest x-ray database and benchmarks on weakly-supervised classification and localization of common thorax diseases},
  author={Wang, Xiaosong and Peng, Yifan and Lu, Le and Lu, Zhiyong and Bagheri, Mohammadhadi and Summers, Ronald M},
  booktitle={Proceedings of the IEEE conference on computer vision and pattern recognition},
  pages={2097--2106},
  year={2017}
}

@inproceedings{irvin2019chexpert,
  title={Chexpert: A large chest radiograph dataset with uncertainty labels and expert comparison},
  author={Irvin, Jeremy and Rajpurkar, Pranav and Ko, Michael and Yu, Yifan and Ciurea-Ilcus, Silviana and Chute, Chris and Marklund, Henrik and Haghgoo, Behzad and Ball, Robyn and Shpanskaya, Katie and others},
  booktitle={Proceedings of the AAAI conference on artificial intelligence},
  volume={33},
  number={01},
  pages={590--597},
  year={2019}
}

@article{qian2024competition,
  title={A competition for the diagnosis of myopic maculopathy by artificial intelligence algorithms},
  author={Qian, Bo and Sheng, Bin and Chen, Hao and Wang, Xiangning and Li, Tingyao and Jin, Yixiao and Guan, Zhouyu and Jiang, Zehua and Wu, Yilan and Wang, Jinyuan and others},
  journal={JAMA ophthalmology},
  volume={142},
  number={11},
  pages={1006--1015},
  year={2024},
  publisher={American Medical Association}
}

@article{decenciere2014feedback,
  title={Feedback on a publicly distributed image database: the Messidor database},
  author={Decenci{\`e}re, Etienne and Zhang, Xiwei and Cazuguel, Guy and Lay, Bruno and Cochener, B{\'e}atrice and Trone, Caroline and Gain, Philippe and Ord{\'o}{\~n}ez-Varela, John-Richard and Massin, Pascale and Erginay, Ali and others},
  journal={Image Analysis \& Stereology},
  pages={231--234},
  year={2014}
}

@article{kriegsmann2022deep,
  title={Deep learning for the detection of anatomical tissue structures and neoplasms of the skin on scanned histopathological tissue sections},
  author={Kriegsmann, Katharina and Lobers, Frithjof and Zgorzelski, Christiane and Kriegsmann, Joerg and Janssen, Charlotte and Meli{\ss}, Rolf R{\"u}dinger and Muley, Thomas and Sack, Ulrich and Steinbuss, Georg and Kriegsmann, Mark},
  journal={Frontiers in Oncology},
  volume={12},
  pages={1022967},
  year={2022},
  publisher={Frontiers Media SA}
}

@article{silva2020going,
  title={Going deeper through the gleason scoring scale: An automatic end-to-end system for histology prostate grading and cribriform pattern detection},
  author={Silva-Rodr{\'\i}guez, Julio and Colomer, Adri{\'a}n and Sales, Mar{\'\i}a A and Molina, Rafael and Naranjo, Valery},
  journal={Computer methods and programs in biomedicine},
  volume={195},
  pages={105637},
  year={2020},
  publisher={Elsevier}
}

@misc{kather2018100,
  title        = {100,000 histological images of human colorectal cancer and healthy tissue (v0.1)},
  author       = {Kather, Jakob Nikolas and Halama, Niels and Marx, Alexander},
  year         = {2018},
  publisher    = {Zenodo},
  doi          = {10.5281/zenodo.1214456},
  url          = {https://doi.org/10.5281/zenodo.1214456}
}

@article{johnson2019mimic,
  title={MIMIC-CXR, a de-identified publicly available database of chest radiographs with free-text reports},
  author={Johnson, Alistair EW and Pollard, Tom J and Berkowitz, Seth J and Greenbaum, Nathaniel R and Lungren, Matthew P and Deng, Chih-ying and Mark, Roger G and Horng, Steven},
  journal={Scientific data},
  volume={6},
  number={1},
  pages={317},
  year={2019},
  publisher={Nature Publishing Group UK London}
}

@inproceedings{zhang2022contrastive,
  title={Contrastive learning of medical visual representations from paired images and text},
  author={Zhang, Yuhao and Jiang, Hang and Miura, Yasuhide and Manning, Christopher D and Langlotz, Curtis P},
  booktitle={Machine learning for healthcare conference},
  pages={2--25},
  year={2022},
  organization={PMLR}
}

@article{silva2025foundation,
  title={A foundation language-image model of the retina (flair): Encoding expert knowledge in text supervision},
  author={Silva-Rodriguez, Julio and Chakor, Hadi and Kobbi, Riadh and Dolz, Jose and Ayed, Ismail Ben},
  journal={Medical Image Analysis},
  volume={99},
  pages={103357},
  year={2025},
  publisher={Elsevier}
}

@article{huang2023visual,
  title={A visual--language foundation model for pathology image analysis using medical twitter},
  author={Huang, Zhi and Bianchi, Federico and Yuksekgonul, Mert and Montine, Thomas J and Zou, James},
  journal={Nature medicine},
  volume={29},
  number={9},
  pages={2307--2316},
  year={2023},
  publisher={Nature Publishing Group US New York}
}

@inproceedings{wang2022medclip,
  title={Medclip: Contrastive learning from unpaired medical images and text},
  author={Wang, Zifeng and Wu, Zhenbang and Agarwal, Dinesh and Sun, Jimeng},
  booktitle={Proceedings of the Conference on Empirical Methods in Natural Language Processing. Conference on Empirical Methods in Natural Language Processing},
  volume={2022},
  pages={3876},
  year={2022}
}

@article{lu2024visual,
  title={A visual-language foundation model for computational pathology},
  author={Lu, Ming Y and Chen, Bowen and Williamson, Drew FK and Chen, Richard J and Liang, Ivy and Ding, Tong and Jaume, Guillaume and Odintsov, Igor and Le, Long Phi and Gerber, Georg and others},
  journal={Nature medicine},
  volume={30},
  number={3},
  pages={863--874},
  year={2024},
  publisher={Nature Publishing Group US New York}
}

@inproceedings{radford2021learning,
  title={Learning transferable visual models from natural language supervision},
  author={Radford, Alec and Kim, Jong Wook and Hallacy, Chris and Ramesh, Aditya and Goh, Gabriel and Agarwal, Sandhini and Sastry, Girish and Askell, Amanda and Mishkin, Pamela and Clark, Jack and others},
  booktitle={International conference on machine learning},
  pages={8748--8763},
  year={2021},
  organization={PMLR}
}
}

\end{document}